\documentclass{article}

    \PassOptionsToPackage{numbers, compress}{natbib}

    \usepackage[preprint]{neurips_2025}

\usepackage{multirow}
\usepackage{makecell}
\usepackage{colortbl}
\usepackage[utf8]{inputenc} 
\usepackage[T1]{fontenc}    
\usepackage{hyperref}       
\usepackage{url}            
\usepackage{booktabs}       
\usepackage{amsfonts}       
\usepackage{nicefrac}       
\usepackage{microtype}      
\usepackage{xcolor}         
\usepackage{graphicx}
\usepackage[most]{tcolorbox}
\usepackage[table]{xcolor}  

\title{BioMedGPT-Mol: Multi-task Learning for Molecular Understanding and Generation}

\author{%
  Chenyang Zuo\textsuperscript{1,2 †}  \,
  Siqi Fan\textsuperscript{1 †} \,
  Zaiqing Nie\textsuperscript{1,2} \thanks{Corresponding author. † Equal contribution. For any questions or discussions, please email \{fansiqi, dair\}@air.tsinghua.edu.cn. }\\
\textsuperscript{1} Institute for AI Industry Research (AIR), Tsinghua University \quad
\textsuperscript{2} PharMolix Inc.\\
}

\begin{document}

\maketitle

\begin{abstract}

  Molecules play a crucial role in biomedical research and discovery, particularly in the field of small molecule drug development. Given the rapid advancements in large language models, especially the recent emergence of reasoning models, it is natural to explore how a general-purpose language model can be efficiently adapted for molecular science applications. In this work, we introduce BioMedGPT-Mol, a molecular language model designed to support molecular understanding and generation tasks. By curating and unifying existing public instruction datasets, we have assembled a large-scale, comprehensive, and high-quality training dataset. The model is then fine-tuned through a meticulously designed multi-task learning framework. On a consolidated benchmark derived from LlaSMol, TOMG-Bench, and MuMOInstruct, BioMedGPT-Mol achieves remarkable performance. Our experimental results demonstrate that a general-purpose reasoning model can be effectively and efficiently post-trained into a professional molecular language model through a well-structured multi-task curriculum. Leveraging these capabilities, we further apply the model to multi-step retrosynthetic planning, achieving state-of-the-art performance on RetroBench and demonstrating its superior efficacy as an end-to-end retrosynthetic planner. We anticipate that our approach can be extended to other biomedical scientific domains.
\end{abstract}

\begin{figure}[h]
\begin{center}
\includegraphics[width=0.95\linewidth]{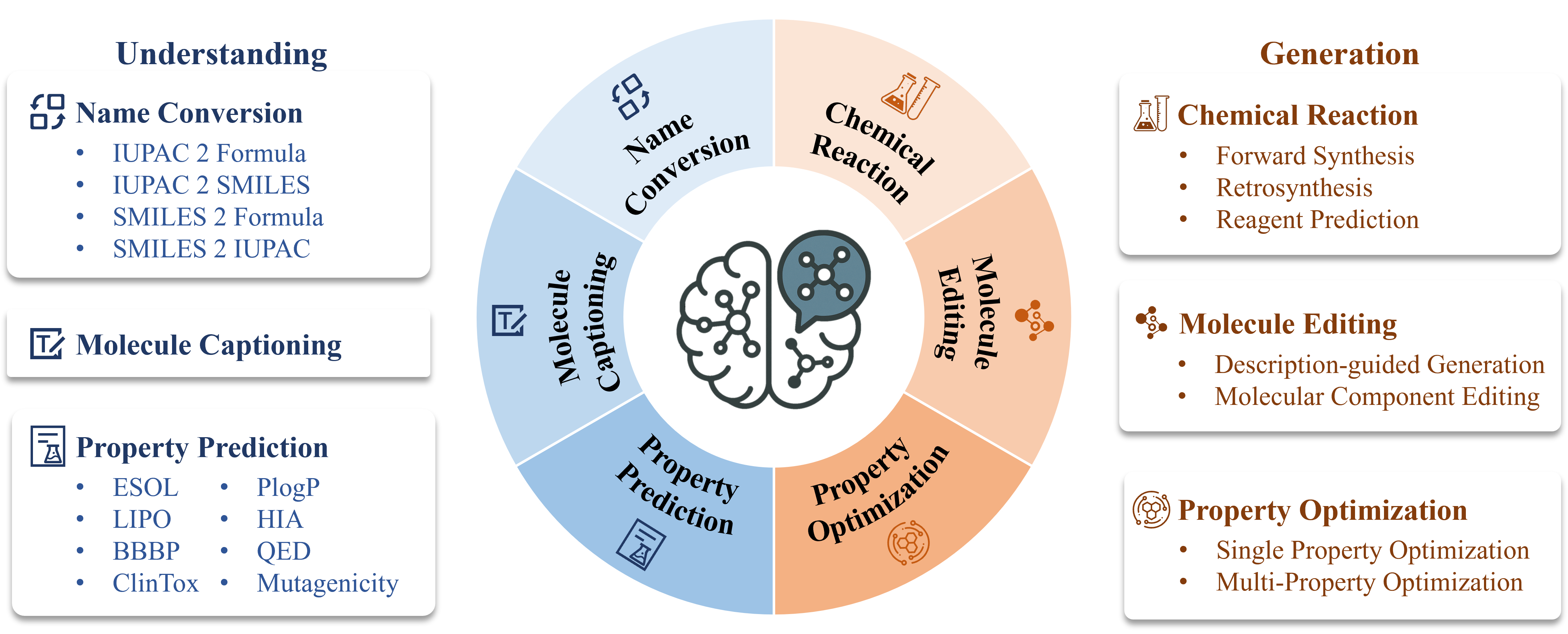}
\end{center}
   \caption{To facilitate molecule-centric scientific discovery, we introduce BioMedGPT-Mol, a molecular language model that simultaneously supports both molecular understanding and generation. Specifically, it is designed to address the following tasks: (1) Molecular understanding tasks, including name conversion, molecule captioning, and property prediction; (2) Molecular generation tasks, encompassing chemical reaction prediction, molecule editing, and property optimization. By integrating these capabilities, BioMedGPT-Mol aims to assist molecular research and development.}
\label{fig:1}
\end{figure}

\section{Introduction}
\label{sec:intro}
Molecules serve as the fundamental tokens of the language of chemistry, and accurate molecular understanding and effective molecule generation are essential for molecule-centric scientific discovery, particularly in the field of small molecule drug development. SMILES (Simplified Molecular Input Line Entry System) \cite{smiles}, a linear molecular notation that encapsulates the structure of chemical compounds, is widely used in both academic papers and patents. This widespread adoption enables large language models (LLMs) to naturally process molecular language. However, the distribution gap between molecular language and natural language necessitates the development of a specialized molecular language model to support molecule-centric research in a question-answering format \cite{guo2023can, chemllm}. Moreover, this molecular language model, as an enhanced base LLM, can further facilitate biomedical multimodal tasks \cite{biomedgpt} and optical chemical structure understanding tasks \cite{ocsu}.

With the rapid advancement of LLMs, general-purpose models have demonstrated their generalization capabilities and can be adapted to various research fields \cite{ds-v3, qwen3}. The development of reasoning models has further shown great potential in scientific discovery \cite{o1, ds-r1, kimi}. Typically, a specialized expert model can be evolved from a general-purpose LLM through in-domain continuous pre-training, specific supervised fine-tuning (SFT), and well-designed reinforcement learning (RL). Although corpora for some research fields have been released \cite{chempile}, the available pre-training data remains finite and limited. Consequently, the focus has shifted to reinforcement learning with verifiable rewards, particularly in the domain of mathematics \cite{deepscaler, llama-nemo}, and more recently in chemistry \cite{ether0}. However, RL post-training remains computationally costly. Given the fast iterative evolution of open-source base LLMs, a natural question arises: Can we efficiently train a biomedical research assistant from a general-purpose reasoning LLM using only supervised fine-tuning? In this work, we explore this possibility in the field of molecular discovery.

Although SFT requires high-quality question-answering data pairs, the learning process is necessary to lead the model to imitate the behaviors of in-domain experts. Several publicly available instruction datasets already exist for molecule-centric scenarios \cite{mol-instructions, llasmol, lm24, tomg-bench, drugassist, mumoinstruct, cotbench}. However, existing works have primarily focused on specific aspects of molecular research, such as molecular understanding \cite{mol-instructions, llasmol}, molecular editing \cite{tomg-bench}, and property optimization \cite{drugassist, tomg-bench, mumoinstruct}. The full potential of reasoning models in the entire molecular discovery journey has yet to be fully explored. 

In this work, we first define the scope of a molecule-centric research assistant for biomedical scientific discovery, particularly for small molecule drug discovery. The molecular language model should address both molecular understanding and generation tasks, as illustrated in Figure~\ref{fig:1}. Thanks to several efforts in dataset construction, most of the required molecular tasks can be supported by existing public datasets, though they are scattered. We curate and unify these datasets to assemble a large-scale, comprehensive, and high-quality training dataset, comprising $13.6$ million question-answering data pairs. To achieve general capability evolution across all relevant tasks, we design a multi-task learning strategy and utilize $5.8$ million data pairs for supervised fine-tuning. This results in BioMedGPT-Mol, a molecular language model tailored for molecular discovery. Specifically, we introduce special tokens to help the model identify molecular language and enable controllable reasoning behaviors for molecular editing and optimization tasks. Finally, we consolidate a benchmark derived from LlaSMol \cite{llasmol}, TOMG-Bench \cite{tomg-bench}, and MuMOInstruct \cite{mumoinstruct} to evaluate the model's general performance in molecular discovery scenarios. BioMedGPT-Mol achieves remarkable performance on this benchmark. The experimental results demonstrate that a general-purpose reasoning model can be effectively and efficiently post-trained into a specialized molecular language model through a well-structured multi-task curriculum. 

Beyond the general capabilities covered by the multi-task framework, we specifically target retrosynthetic planning, a fundamental problem in organic chemistry aiming to identify a set of commercially available starting materials and a sequence of reactions that can synthesize a given target molecule. To the best of our knowledge, LLMs alone have not previously been deployed as effective multi-step retrosynthetic planners. In this work, we further advance BioMedGPT-Mol to tackle this challenge using a comprehensive training pipeline integrating SFT and RL. Consequently, our model achieves state-of-the-art performance on RetroBench \cite{liu2023fusionretro}, demonstrating its superior capability of acting as a fully end-to-end retrosynthetic planner.

We anticipate that our approach can be extended to other biomedical scientific domains to leverage the power of open-source reasoning models.
Our contributions are summarized as follows:
\begin{itemize}
  \item We define the scope of a molecule-centric research assistant for biomedical scientific discovery and introduce BioMedGPT-Mol, a molecular language model designed to advance molecular discovery.
  \item We present a multi-task learning strategy to enable the model to both understand and generate molecular language. The designed special tokens facilitate controllable reasoning behaviors for molecular editing and optimization tasks. Additionally, we assemble a large-scale, comprehensive, and high-quality dataset for training and consolidate a benchmark for evaluation.
  \item We validate the effectiveness of the proposed approaches through comprehensive experiments and demonstrate the performance advantages of BioMedGPT-Mol in the context of molecular discovery tasks.
  \item We are the first to explore the retrosynthetic planning task using an LLM alone. We introduce a comprehensive training pipeline integrating SFT and RL, enabling BioMedGPT-Mol to achieve state-of-the-art performance as an end-to-end retrosynthetic planner.
\end{itemize}

\section{Related Work}

Since the advent of large language models, they have been widely employed in chemistry applications\cite{guo2023can, white2023assessment, jablonka2024leveraging}. Several recent efforts have focused on either fine-tuning LLMs with new datasets or conducting benchmark studies \cite{llasmol}. ChemLLM \cite{chemllm} introduces a framework featuring the first LLM dedicated to chemistry, including a dataset, ChemData, for training and a benchmark, ChemBench, for evaluation. Mol-Instructions \cite{mol-instructions} covers a broader range of biomolecular tasks and releases a larger instruction dataset. LlaSMol \cite{llasmol} is fine-tuned on SMolInstruct, a larger and more comprehensive dataset, achieving performance improvements. Molecule discovery involves not only understanding existing chemical compounds but also innovating and generating new ones. TOMG-Bench \cite{tomg-bench} evaluates LLMs on text-based open molecule generation and constructs an instruction tuning dataset, OpenMolIns. DrugAssistant \cite{drugassist} focuses on property optimization, while GeLLM3O \cite{mumoinstruct} explores joint optimization for multiple properties. In contrast, ChemPile \cite{chempile} presents a large-scale dataset for pretraining chemistry foundation models. Additionally, ether0 \cite{ether0} is a chemistry reasoning model fine-tuned through reinforcement learning, and ChemCoTDataset \cite{cotbench} is released to enhance chemistry reasoning. The existing datasets are scattered, and the full potential of fine-tuned reasoning models in the entire molecular discovery journey has yet to be fully explored. We curate and unify these datasets to enhance general capabilities across all relevant molecular discovery tasks, resulting in BioMedGPT-Mol, a new molecular language model.

\section{Problem Formulation}
Accurate molecular understanding and effective molecule generation are crucial for molecule-centric scientific discovery. To develop a LLM-based research assistant that can facilitate molecular research, we first delineate the task scope, spanning from molecular understanding to generation.

\subsection{Molecular Understanding Tasks}
\subsubsection{Name Conversion}
Molecular name serves as the fundamental knowledge for molecular language. Name conversion \cite{mol-instructions, llasmol} requires the model to convert the name of a molecule from one representation to another, including converting IUPAC name to SMILES (I2S), converting IUPAC name to molecular formula (I2F), converting SMILES to IUPAC (S2I), and converting SMILES to molecular formula (S2F). 

\begin{tcolorbox}[colback=gray!10!white,colframe=lightgray,title=Example of Name Conversion Task - IUPAC 2 SMILES]
\textbf{Query}: Can you give the SMILES notation of the molecule 4-N-(6-chloro-2-methoxyacridin-9-yl)-1-N,1-N-diethylpentane-1,4-diamine ?

\textbf{Answer}: Sure. CCN(CC)CCCC(C)NC1=C2C=CC(Cl)=CC2=NC2=CC=C(OC)C=C12
\label{task: name_conversion_i2s}
\end{tcolorbox}

\begin{tcolorbox}[colback=gray!10!white,colframe=lightgray,title=Example of Name Conversion Task - IUPAC 2 Formula]
\textbf{Query}: Translate the chemical IUPAC name 4-N-(6-chloro-2-methoxyacridin-9-yl)-1-N,1-N-diethylpentane-1,4-diamine into its molecular formula.

\textbf{Answer}: C23H30ClN3O
\label{task: name_conversion_i2f}
\end{tcolorbox}

\begin{tcolorbox}[colback=gray!10!white,colframe=lightgray,title=Example of Name Conversion Task - SMILES 2 IUPAC]
\textbf{Query}: Can you give the IUPAC name of the molecule CCN(CC)CCCC(C)NC1=C2C= CC(Cl)=CC2=NC2=CC=C(OC)C=C12?

\textbf{Answer}: Sure. 4-N-(6-chloro-2-methoxyacridin-9-yl)-1-N,1-N-diethylpentane-1,4-diamine.
\label{task: name_conversion_s2i}
\end{tcolorbox}

\begin{tcolorbox}[colback=gray!10!white,colframe=lightgray,title=Example of Name Conversion Task - SMILES 2 Formula]
\textbf{Query}: I'd like to know the molecular formula of CCN(CC)CCCC(C)NC1=C2C=CC (Cl)=CC2=NC2=CC=C(OC)C=C12. Can you tell me?

\textbf{Answer}: Sure. It's C23H30ClN3O.
\label{task: name_conversion_s2f}
\end{tcolorbox}

\subsubsection{Molecule Captioning}
Molecule captioning \cite{mol-instructions, llasmol, lm24} involves generating a detailed textual description that elucidates the structure, properties, biological activity, and applications of a molecule based on its SMILES. By providing chemists and biologists with a quick and accessible summary of essential molecular information, it can efficiently guide their research and experimental efforts.

\begin{tcolorbox}[colback=gray!10!white,colframe=lightgray,title=Example of Molecule Captioning Task]
\textbf{Query}: I'd like a short overview about this molecule. Can you do that? CCCCCCCCCCC CCCCCCCCCCCCCC(=O)O[C@H](COC(=O)CCCCCCCCCCCCCCCCCCC)COP(=O) (O)OC[C@@H](O)COP(=O)(O)OC[C@@H](COC(=O)CCCCCCCCC(C)CC)OC(=O)C CCCCCCCCCCC(C)C

\textbf{Answer}: The molecule is a stabilizing mitochondrial structure, proton trap for oxidative phosphorylation, cholesterol translocation that impacts barth syndrome and tangier disease. The molecule is a stabilizing cytochrome oxidase and a apoptosis that impacts aging, diabetic heart disease, and non-alcoholic fatty liver disease.
\label{task: mol_cap}
\end{tcolorbox}

\subsubsection{Property Prediction}
Property prediction \cite{moleculenet, mol-instructions, llasmol} involves estimating a molecule's inherent properties based on structural characteristics derived from its SMILES. By enabling the prediction of unknown attributes of new molecules, it may enhance research efficiency and reduces development times. It includes several subtasks.

plogP represents lipophilicity, balancing permeability, solubility, and metabolic stability. ESOL prediction \cite{esol} estimates water solubility, while Lipo prediction \cite{lipo} assesses the octanol/water distribution coefficient, an important feature that influences both membrane permeability and solubility. Human Intestinal Absorption (HIA) prediction evaluates a molecule’s ability to be absorbed through the gastrointestinal tract.

\begin{tcolorbox}[colback=gray!10!white,colframe=lightgray,title=Example of Property Prediction - plogP]
\textbf{Query}: Please predict the penalized octanol-water partition coefficient of [NH3+]C(CC1=c 2cccc(O)c2=[NH+]C1)C(=O)[O-].

\textbf{Answer}: The penalized octanol-water partition coefficient is -7.84.
\label{task: pp_plogP}
\end{tcolorbox}

\begin{tcolorbox}[colback=gray!10!white,colframe=lightgray,title=Example of Property Prediction - ESOL]
\textbf{Query}: What is the logarithmic value of the solubility of CC\#N in water?

\textbf{Answer}: 0.26 mol/L.
\label{task: pp_esol}
\end{tcolorbox}

\begin{tcolorbox}[colback=gray!10!white,colframe=lightgray,title=Example of Property Prediction - Lipo]
\textbf{Query}: Provide the logD (the octanol/water distribution coefficient) at pH 7.4 of O=C C1=CC=CC=C1B(O)O.

\textbf{Answer}: 0.83.
\label{task: pp_lipo}
\end{tcolorbox}

\begin{tcolorbox}[colback=gray!10!white,colframe=lightgray,title=Example of Property Prediction - HIA]
\textbf{Query}: Please predict the intestinal adsorption of [NH3+]C(CC1=c2cccc(O)c2=[NH+] C1)C(=O)[O-].

\textbf{Answer}: The intestinal adsorption is 0.7.
\label{task: pp_hia}
\end{tcolorbox}

The blood-brain barrier (BBB), which separates circulating blood from brain extracellular fluid, blocks most drugs, hormones, and neurotransmitters. This barrier presents a long-standing challenge in the development of drugs targeting the central nervous system (CNS). Blood-brain barrier penetration (BBBP) prediction involves two tasks: a classification task \cite{bbbp} to determine whether a molecule can cross the BBB, and a regression task to estimate the molecule's ability to cross. For CNS drug candidates, higher BBBP is desirable.

\begin{tcolorbox}[colback=gray!10!white,colframe=lightgray,title=Example of Property Prediction - BBBP]
\textbf{Query}: Is CNC(C)C(=O)C1=CC=C(OC)C=C1 capable of crossing the blood-brain barrier?

\textbf{Answer}: No.

\textbf{Query}: Please predict the BBB permeability of [NH3+]C(CC1=c2cccc(O)c2=[NH+] C1)C(=O)[O-].

\textbf{Answer}: The BBB permeability is 0.27.
\label{task: pp_bbbp}
\end{tcolorbox}

QED (Quantitative Estimate of Drug-likeness) prediction is to predict the overall drug-likeness of a molecule by integrating multiple molecular attributes, including molecular weight, lipophilicity, and solubility.

\begin{tcolorbox}[colback=gray!10!white,colframe=lightgray,title=Example of Property Prediction - QED]
\textbf{Query}: Please predict the QED of [NH3+]C(CC1=c2cccc(O)c2=[NH+]C1)C(=O)[O-].

\textbf{Answer}: The QED is 0.48.
\label{task: pp_qed}
\end{tcolorbox}

ClinTox prediction \cite{clintox} is to predict whether the molecule is toxic to human body.

\begin{tcolorbox}[colback=gray!10!white,colframe=lightgray,title=Example of Property Prediction - ClinTox]
\textbf{Query}: Does CC(C)[NH2+]C(C)CC1=CC=C(I)C=C1 have toxicity?

\textbf{Answer}: No.
\label{task: pp_clintox}
\end{tcolorbox}

Mutagenicity (Mutag) prediction involves forecasting the likelihood that a drug will induce genetic mutations.

\begin{tcolorbox}[colback=gray!10!white,colframe=lightgray,title=Example of Property Prediction - Mutag]
\textbf{Query}: Predict the mutagenicity of [NH3+]C(CC1=c2cccc(O)c2=[NH+]C1)C(=O)[O-].

\textbf{Answer}: The mutagenicity is 0.2.
\label{task: pp_mutag}
\end{tcolorbox}

\subsection{Molecular Generation Tasks}

\subsubsection{Chemical Reaction}
The chemical reaction prediction task encompasses three subtasks: forward synthesis, retrosynthesis, and reagent prediction.

Forward reaction prediction \cite{mol-instructions, llasmol} aims to identify potential products given specific reactants and reagents. This subtask streamlines research and development processes reduces experimental guesswork.

\begin{tcolorbox}[colback=gray!10!white,colframe=lightgray,title=Example of Chemical Reaction - Forward Synthesis]
\textbf{Query}: C1CCOC1. CCN(CC)CC. CS(=O)(=O)Cl. CS(C)=O. N[C@@H]1CC2=CC=C (CN3C=C(CO)C(C(F)(F)F)=N3)C=C2C1 Considering the given starting materials, what might be the resulting product in a chemical reaction?

\textbf{Answer}: CS(=O)(=O)N[C@@H]1CC2=CC=C(CN3C=C(CO)C(C(F)(F)F)=N3)C=C2C1
\label{task: cr_fs}
\end{tcolorbox}

Retrosynthetic analysis \cite{mol-instructions, llasmol} predicts potential reactants given a target product. This approach is crucial for designing efficient synthetic strategies for complex molecules, thereby accelerating the development of novel pharmaceuticals.

\begin{tcolorbox}[colback=gray!10!white,colframe=lightgray,title=Example of Chemical Reaction - Retrosynthesis]
\textbf{Query}: Based on the given product, provide some plausible reactants that might have been utilized to prepare it. N\#CCC1=CC(Cl)=CC(CC2=NC3=C(F)C(F)=CC(F)=C3S2)=C1

\textbf{Answer}: N\#CCC1=CC(Cl)=CC(CC\#N)=C1.NC1=C(F)C(F)=CC(F)=C1S
\label{task: cr_rs}
\end{tcolorbox}

Reagent prediction \cite{mol-instructions} focuses on identifying the appropriate catalysts, solvents, or auxiliary substances required for a specific chemical reaction. By doing so, it helps chemists optimize reaction conditions, ultimately enhancing the overall efficiency and cost-effectiveness of chemical processes.

\begin{tcolorbox}[colback=gray!10!white,colframe=lightgray,title=Example of Chemical Reaction - Reagent Prediction]
\textbf{Query}: Based on the given chemical reaction, can you propose some likely reagents that might have been utilized? O=CC1=CC([N+1](=O)[O-1])=C(F)C=C1F >> O=[N+1]([O-1])C1=CC(CO)=C(F)C=C1F

\textbf{Answer}: C1CCOC1. Cl. [BH4-1]. [Na+1]
\label{task: cr_rp}
\end{tcolorbox}

\subsubsection{Molecule Editing}
Molecule editing involves generating a new molecule that either meets a specific description or follows an explicit modification instruction. This task further comprises two main subtasks: description-guided generation and molecular component editing.

Description-guided molecule generation \cite{mol-instructions, llasmol} has the potential to streamline the process of molecular design by enabling the production of molecules that directly meet the criteria outlined in a given description. This approach facilitates a more targeted method for creating novel molecules, with applications in diverse fields such as drug discovery.

\begin{tcolorbox}[colback=gray!10!white,colframe=lightgray,title=Example of Molecule Editing - Description-guided Generation]
\textbf{Query}: Can you create a molecule that matches the given characteristics? The molecule is a branched amino pentasaccharide consisting of D-glucose at the reducing end having an alpha-L-fucosyl-(1->3)-[beta-D-galactosyl-(1->4)]-N-acetyl-beta-D-glucosaminyl-(1->3)-beta-D-galactosyl moiety attached at the 4-position. It is an amino pentasaccharide and a glucosamine oligosaccharide.

\textbf{Answer}: Sure. CC(=O)N[C@H]1[C@H](O[C@H]2[C@@H](O)[C@@H](CO)O[C@@H] (O[C@H]3[C@H](O)[C@@H](O)C(O)O[C@@H]3CO)[C@@H]2O)O[C@H](CO)[C@ @H](O[C@@H]2O[C@H](CO)[C@H](O)[C@H](O)[C@H]2O)[C@@H]1O[C@@H]1O [C@@H](C)[C@@H](O)[C@@H](O)[C@@H]1O.
\label{task: mol_edit_dgg}
\end{tcolorbox}

Molecular component editing \cite{tomg-bench} is more straightforward and requires the model to make modifications to a given molecule, leveraging its knowledge of molecular structure. Specifically, the model is challenged to: (1) Add or remove a specific functional group, (2) Remove a designated functional group and then introduce a new one as specified to the molecule.

\begin{tcolorbox}[colback=gray!10!white,colframe=lightgray,title=Example of Molecule Editing - Molecular Component Editing]
\textbf{Query}: Add a hydroxyl to N\#CC(=NNc1ccccc1)c1nc(-c2ccc3ccccc3c2)c(N=Nc2ccccc2)s1.

\textbf{Answer}: N\#CC(=NNc1cccc(O)c1)c1nc(-c2ccc3ccccc3c2)c(N=Nc2ccccc2)s1

\textbf{Query}: Modify CC(C)Oc1cncc(NCc2c(C(F)(F)F)cnn2C)n1 by removing a halo.

\textbf{Answer}: CC(C)Oc1cncc(NCc2c(C(F)F)cnn2C)n1

\textbf{Query}: Please substitute a nitrile in COc1cccc2c(C)cc(SCC\#N)nc12 with a halo.

\textbf{Answer}: COc1cccc2c(C)cc(SCI)nc12

\label{task: mol_edit_mce}
\end{tcolorbox}

\subsubsection{Property Optimization}

Property optimization tasks instruct the model to not only edit molecules but also to discern whether the modifications will steer the molecule toward a desired optimization target.

First, the model is tasked with optimizing a single property, such as logP, BBBP, MR (molecular refractivity), or QED, which pertains to the potential pharmacological attributes of the molecule.

\begin{tcolorbox}[colback=gray!10!white,colframe=lightgray,title=Example of Property Optimization - Single Property Optimization]
\textbf{Query}: Please modify the molecule NC(=[NH2+])c1ccc(N2CCC(c3ccccc3)C2)nc1 to increase its QED value.

\textbf{Answer}: NCc1ccc(N2CCC(c3ccccc3)C2)nc1.
\label{task: prop_opt_single}
\end{tcolorbox}

Then, the model is challenged to jointly optimize multiple properties, which represent realistic challenges in lead optimization for drug development. Several representative properties are selected, including plogP, QED, BBBP, mutagenicity, and HIA.

\begin{tcolorbox}[colback=gray!10!white,colframe=lightgray,title=Example of Property Optimization - Multi-Property Optimization]
\textbf{Query}: Modify the molecule [NH3+]C(CC1=c2cccc(O)c2=[NH+]C1)C(=O)[O-] to increase its human intestinal adsorption ability value, decrease its mutagenicity value, increase its penalized octanol-water partition coefficient value, and increase its drug-likeness quantified by QED score value. Keep the modifications to the molecule structure as minimal as possible.

\textbf{Answer}: O=C([O-])C(CC1=c2cccc(O)c2=[NH+]C1)c1cccc(Br)c1
\label{task: prop_opt_multi}
\end{tcolorbox}

\section{Methodology}
Molecular understanding and generation can be regarded as the translation between molecular language and natural language. To enhance this process, we first introduce special tokens to help the model identify molecular language. We then employ multi-task learning to encourage the model to capture the intrinsic distribution and enhance the generalization ability. Finally, we explore to enable controllable reasoning behaviors for molecular reasoning tasks, including editing and optimization.

\subsection{Special Tokens for Molecular Language}
We adopt several widely used special tokens to help the model identify the start and end of molecular language, following previous works \cite{mol-instructions, llasmol}. Specifically, there are three typical naming paradigms for a molecule: SMILES strings, IUPAC name, and molecular formula. These paradigms form the foundation of molecular language, and we introduce three corresponding tags to distinguish them.

\begin{itemize}
  \item The \texttt{<SMILES>} \texttt{</SMILES>} tag is used to identify SMILES strings. It not only helps the model distinguish SMILES strings but also facilitates the parsing of generated molecules in molecular generation tasks.
  \item The \texttt{<IUPAC>} \texttt{</IUPAC>} tag is used to identify IUPAC names. 
  \item The \texttt{<MOLFORMULA>} \texttt{</MOLFORMULA>} tag is used to identify molecular formula strings.
\end{itemize}

We preprocess the training data pairs to ensure that all SMILES strings, IUPAC names, and molecular formulas are enclosed within their corresponding tags.

\subsection{Multi-task Learning Strategy}

\begin{figure}[ht]
\begin{center}
\includegraphics[width=0.95\linewidth]{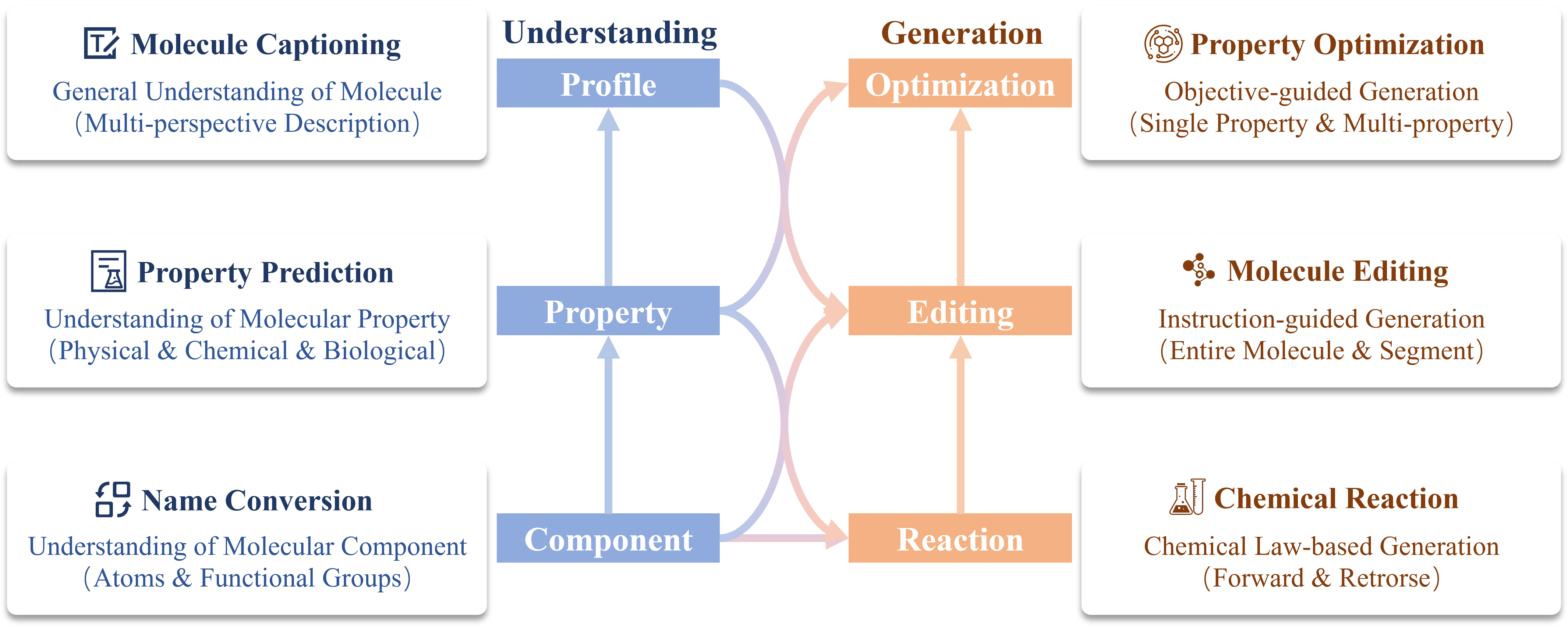}
\end{center}
   \caption{The various molecular tasks are interconnected, and multi-task learning enables molecular language models to capture the intrinsic essential attributes. Molecular understanding evolves from the concrete components of a molecule to its specific properties, and ultimately to an abstract, generalized description. Similarly, the guidance for molecule generation progresses from intrinsic chemical laws to detailed instructions, and then to specific property objectives. Accurate molecular understanding forms the foundation for effective molecule generation. In essence, better understanding leads to better generation.}
\label{fig:2}
\end{figure}

Although molecular understanding and generation tasks are diverse, they are interconnected and exhibit dependencies, as illustrated in Figure~\ref{fig:2}.

SMILES strings, IUPAC names, and molecular formulas are the three typical naming paradigms in molecular language, each describing a molecule from a unique perspective. Molecular formulas represent a molecule at the atomic level, specifying the number of each type of atom present. SMILES strings provide additional structural information, organizing atoms according to the molecular structure, including functional groups. IUPAC names, which are readable by chemists, further introduce semantic information by incorporating the natural language names of functional groups. The name conversion task, as a foundational task, enables the model to understand molecular components from the atomic level to the structural level, and ultimately to the semantic level.

Building on the atomic components and structural characteristics derived from SMILES, the model is required to predict specific inherent properties, including physical, chemical, and biological properties. This challenges the model to capture the relationship between structure and property. The model is then asked to provide a profile based on its understanding of molecular components and properties.

With an accurate understanding of existing molecules, the model is then challenged to learn to generate new ones. First, the chemical reaction prediction task requires the model to simulate chemical laws, leveraging its knowledge of molecular structure and properties. Next, the model is required to generate a molecule or modify segments of a given molecule based on detailed instructions. The model must understand both the composition of the molecule and the intention of the instructions. Finally, the model is encouraged to edit a given molecule to meet a specific property objective. This requires the model to predict the specific property of the source molecule, understand the optimization intention, and perform appropriate edits.

We employ a multi-task learning strategy to encourage the molecular language model to capture the intrinsic distribution and enhance its generalization ability for molecule-centric discovery.

\subsection{A Large-scale, Comprehensive, and High-quality Dataset}

To enable multi-task learning, we first review, curate, and unify existing molecule-related datasets, resulting in a large-scale, comprehensive, and high-quality dataset. 

We adopt SMILES as the general molecular representation in all tasks except for name conversion, following the findings in \cite{llasmol}. During pre-processing, we enclose all SMILES strings, IUPAC names, and molecular formulas within their corresponding tags. To facilitate the property optimization task, we introduce auxiliary property prediction tasks, including plogP, QED, BBBP, Mutagenicity, and HIA regression tasks. The data pairs for these tasks are generated based on the raw property data provided in MuMOInstruct \cite{mumoinstruct}.

A detailed description of the reconstructed dataset is provided in Table~\ref{tab:data}.

\begin{table}[ht]
    \centering
    \caption{\textbf{Statistics of the Reconstructed Dataset.} cls: classification task. reg: regression task.}
    \begin{tabular}{p{1.5cm}p{1.7cm}|c c p{3cm}}
    \hline
    \textbf{Category}            & \textbf{Subcategory}       & \textbf{Name}    & \textbf{\# K Pairs}   & \textbf{Sources}   \\ \hline
    \multirow{17}{*}{\makecell[c]{Underst-\\anding}}& \multirow{4}{*}{\makecell[c]{Name\\Conversion}} & I2S & 299.9 & SMolInstruct \cite{llasmol} \\
                                 &                                                                  & I2F & 299.9 & SMolInstruct \cite{llasmol} \\
                                 &                                                                  & S2I & 299.9 & SMolInstruct \cite{llasmol} \\
                                 &                                                                  & S2F & 299.9 & SMolInstruct \cite{llasmol} \\ \cline{2-5}
                                 & \multirow{2}{*}{\makecell[c]{Molecule\\Captioning}}  & \multirow{2}{*}{\makecell[c]{Molecule Captioning}} & \multirow{2}{*}{\makecell[tl]{205.7}} & SMolInstruct \cite{llasmol} \\ 
                                 & & & & L+M-24 \cite{lm24} \\ \cline{2-5}
                                 & \multirow{11}*{\makecell[c]{Property\\Prediction}} & ClinTox-cls & 1.1 & SMolInstruct \cite{llasmol} \\
                                 &                                                    & BBBP-cls & 1.6 & SMolInstruct \cite{llasmol} \\
                                 &                                                    & HIV-cls     & 32.9 & SMolInstruct \cite{llasmol} \\
                                 &                                                    & Sider-cls   & 1.1  & SMolInstruct \cite{llasmol} \\
                                 &                                                    & ESOL-reg    & 0.9  & SMolInstruct \cite{llasmol} \\
                                 &                                                    & LIPO-reg    & 3.4  & SMolInstruct \cite{llasmol} \\ \cline{5-5}
                                 &            & plogP-reg & 1656.2 & \multirow{5}*{\makecell[c]{Constructed\\based on\\MuMOInstruct \cite{mumoinstruct}}} \\
                                 &            & QED-reg   & 1656.2 & \\
                                 &            & BBBP-reg  & 1656.2 & \\
                                 &            & Mutag-reg & 1656.2 & \\
                                 &            & HIA-reg   & 1656.2 & \\ \hline
    \multirow{7}*{\makecell[c]{Generation}}   & \multirow{3}*{\makecell[c]{Chemical\\Reaction}} & Forward Synthesis & 971.8 & SMolInstruct \cite{llasmol} \\
                                 &                                                               & Retrosynthesis    & 941.7 & SMolInstruct \cite{llasmol} \\
                                 &                                                               & Reagent Prediction & 125.4 & Mol-Instructions \cite{mol-instructions} \\ \cline{2-5}
                                 & \multirow{2}*{\makecell[c]{Molecule\\Editing}} & Description-guided Generation & 56.5 & SMolInstruct \cite{llasmol} \\
                                 &                                                & Molecule Component Editing & 55.7 & OpenMolIns \cite{tomg-bench} \\ \cline{2-5}
                    & \multirow{2}*{\makecell[c]{Property\\Optimization}} & Single Property Optimization & 885.9 & MolOpt-Instruct \cite{drugassist} \\
                    &                                                     & Multi-property Optimization & 828.1 & MuMOInstruct \cite{mumoinstruct} \\ \hline
    \multicolumn{3}{c}{\textbf{Total}} & \multicolumn{2}{|c}{\textbf{$13,592.4$ K $\approx 13.6$ M}} \\ \hline

    \end{tabular}
    \label{tab:data}
\end{table}

\subsection{Controllable Behaviors for Molecular Reasoning}
Although explicit reasoning (contents within the `\texttt{<think>}' tag) makes the model's behavior more explainable, it also introduces additional computational cost. In fact, not every task requires the inclusion of a `thinking' process, even in molecular reasoning tasks. To enable users to manage and control reasoning behavior, we add `thinking' data to molecular reasoning tasks, including molecular editing and property optimization. Inspired by Qwen3 \cite{qwen3}, we adopt a `\texttt{\/no\_think}' flag to help the model recognize when to follow `non-thinking' behavior.

\begin{tcolorbox}[colback=gray!10!white,colframe=lightgray,title=Example of Molecular Editing in Thinking Behavior]
\textbf{Query}: Add a carboxyl to the molecule CC1[NH2+]CCC1C(=O)Nc1cc(C(N)=O)ccc1Cl.

\textbf{Answer}: \texttt{<think>}
\begin{itemize}
    \item Molecule Analysis: The input molecule contains a cyclohexane ring with an ammonium group ([NH2+]), a ketone (C=O), and an amide (C(=O)N) linked to a substituted benzene ring. The benzene ring has a urea-like group (C(N)=O) and a chlorine substituent. The primary functional groups are ammonium, ketone, amide, and urea.
    \item Editing Strategy: The benzene ring has available positions for substitution. The chlorine and urea-like groups are meta/para-directing. Adding a carboxyl group (-COOH) is most feasible at the meta position relative to the urea-like group (C(N)=O) to align with aromatic substitution patterns. This avoids steric hindrance and maintains electronic compatibility.
    \item Feasibility Analysis: While the benzene ring is deactivated by electron-withdrawing groups (Cl, C(N)=O), carboxylation via directed ortho/metalation or coupling reactions could achieve this modification. The proposed site (meta to C(N)=O) is chemically reasonable, though harsh conditions or specialized reagents may be required.
\end{itemize}
\texttt{</think>}
CC1[NH2+]CCC1C(=O)Nc1cc(C(N)=O)cc(C(=O)O)c1Cl
\label{task: cot_edit}
\end{tcolorbox}

\begin{tcolorbox}[colback=gray!10!white,colframe=lightgray,title=Example of Single Property Optimization in Thinking Behavior]
\textbf{Query}: Can you optimize the molecule C[N+]([O-])=Cc1cccc(Oc2ccccc2)c1 to improve its drug-likeness?

\textbf{Answer}: \texttt{<think>}
\begin{itemize}
    \item Structural Analysis of the Molecule: The source molecule contains a nitro group ([N+]([O-])=C) attached to a methylene group, which is connected to a substituted benzene ring (diphenyl ether scaffold). The nitro group is electron-withdrawing and may contribute to high lipophilicity and metabolic instability.
    \item Property Analysis: The nitro group increases molecular weight, logP, and potential toxicity, reducing QED (drug-likeness). It also introduces a strong electron-withdrawing effect, which may hinder solubility and metabolic stability.
    \item Limitation in the Molecule for Property: The nitro group is a key limitation due to its negative impact on QED parameters (e.g., high logP, metabolic liability, and potential genotoxicity).
    \item Optimization for the Molecule: Replace the nitro group (-NO2) with a sulfonate ester (-SO3R), a bioisostere that reduces lipophilicity (lower logP), improves solubility, and eliminates nitro-related toxicity. The diphenyl ether scaffold is retained for structural integrity.
\end{itemize}
\texttt{</think>}
CCS(=O)(=O)Oc1cccc(Oc2ccccc2)c1
\label{task: cot_opt}
\end{tcolorbox}

To achieve this, we generate CoT (Chain of Thought) data pairs based on the raw data from ChemCoTDataset \cite{cotbench}, which imitates experts' reasoning behaviors. For `non-thinking' behavior, an empty thinking block is retained in the assistant's response to ensure internal format consistency. Similar to Qwen3, the model operates in thinking behavior by default. Examples of molecular reasoning tasks in `thinking' behavior are presented above. The statistics of the molecular CoT dataset are provided in Table~\ref{tab:data_cot}. 

\begin{table}[ht]
    \centering
    \caption{\textbf{Statistics of Molecular CoT Dataset.} cls: classification task. reg: regression task.}
    \begin{tabular}{ccc}
    \hline
    \textbf{Category}            & \textbf{Name}   & \textbf{\# K Pairs}   \\ \hline
    \multirow{2}*{\makecell[c]{CoT}}  & Molecular Editing & 4.5 \\
                                      & Single Property Optimization & 2.8 \\ \hline
    \multicolumn{2}{c}{Total} & 7.3 \\ \hline                            
    \end{tabular}
    \label{tab:data_cot}
\end{table}

\subsection{Implementation of BioMedGPT-Mol}

We first construct the dataset for model fine-tuning, utilizing most of the data reported in Table~\ref{tab:data}. For the auxiliary property prediction tasks, we randomly sample $100,000$ data pairs for each task. The instructions of all these data pairs are concatenated with the `\texttt{\/no\_think}' flag. Additionally, the generated CoT data pairs are integrated. In total, the entire training set comprises approximately $5.8$ million question-answering pairs.

We initialize BioMedGPT-Mol with the pre-trained parameters of Qwen3-8B \cite{qwen3}. The model is fine-tuned for 3 epochs with a learning rate of $1 \times 10^{-4}$ and a equivalent batch size of $256$. To reduce training costs, we apply low-rank adaptation (LoRA) to BioMedGPT-Mol. Additionally, we perform linear warmup for the first $10\%$ of steps and use a cosine annealing strategy to stabilize training. The entire training process requires approximately $624$ GPU hours on Nvidia A800 GPUs.

\section{Description of Benchmark}

To evaluate the molecular language model's general performance in molecular discovery scenarios, we consolidate a benchmark derived from LlaSMol \cite{llasmol}, TOMG-Bench \cite{tomg-bench}, and MuMOInstruct \cite{mumoinstruct}.

\subsection{Tasks}
We select $58,904$ question-answering pairs across $19$ representative tasks for practical applications such as small molecule drug discovery, covering a spectrum from molecular understanding to generation. The statistical information is reported in Table~\ref{tab:bench}.

\begin{table}[ht]
    \centering
    \caption{\textbf{Statistics of the Consolidated Benchmark.} cls: classification task. reg: regression task.}
    \begin{tabular}{p{1.5cm}p{1.7cm}|c c p{3cm}}
    \hline
    \textbf{Category}            & \textbf{Subcategory}       & \textbf{Name}    & \textbf{\# Pairs}   & \textbf{Sources}   \\ \hline
    \multirow{10}{*}{\makecell[c]{Underst-\\anding}}& \multirow{4}{*}{\makecell[c]{Name\\Conversion}} & I2S & 2,993 & SMolInstruct \cite{llasmol} \\
                                 &                                                                  & I2F & 2,993 & SMolInstruct \cite{llasmol} \\
                                 &                                                                  & S2I & 2,993 & SMolInstruct \cite{llasmol} \\
                                 &                                                                  & S2F & 2,993 & SMolInstruct \cite{llasmol} \\ \cline{2-5}
                                 & \multirow{2}{*}{\makecell[c]{Molecule\\Captioning}}  & \multirow{2}{*}{\makecell[c]{Molecule Captioning}} & \multirow{2}{*}{\makecell[tl]{2,538}} & SMolInstruct \cite{llasmol} \\ 
                                 & & & & \\ \cline{2-5}
                                 & \multirow{4}*{\makecell[c]{Property\\Prediction}} & ClinTox-cls & 144 & SMolInstruct \cite{llasmol} \\
                                 &                                                    & BBBP-cls & 197 & SMolInstruct \cite{llasmol} \\
                                 &                                                    & ESOL-reg    & 112  & SMolInstruct \cite{llasmol} \\
                                 &                                                    & LIPO-reg    & 420  & SMolInstruct \cite{llasmol} \\ \hline
    \multirow{10}*{\makecell[c]{Generation}}   & \multirow{2}*{\makecell[c]{Chemical\\Reaction}} & Forward Synthesis & 4,062 & SMolInstruct \cite{llasmol} \\
                                 &                                                              & Retrosynthesis    & 4,156 & SMolInstruct \cite{llasmol} \\ \cline{2-5}
                                 & \multirow{4}*{\makecell[c]{Molecule\\Editing}} & Add Component & 5,000 & TOMG-Bench \cite{tomg-bench} \\
                                 &                                                & Delete Component & 5,000 & TOMG-Bench \cite{tomg-bench} \\
                                 &                                                & Substitute Component & 5,000 & TOMG-Bench \cite{tomg-bench} \\
                                 &                                                & Description-guided Generation & 2,493  & SMolInstruct \cite{llasmol} \\ \cline{2-5}
                    & \multirow{4}*{\makecell[c]{Property\\Optimization}} & logP Optimization & 5,000 & TOMG-Bench \cite{tomg-bench} \\
                    &                                                     & QED Optimization & 5,000 & TOMG-Bench \cite{tomg-bench} \\
                    &                                                     & MR Optimization & 5,000 & TOMG-Bench \cite{tomg-bench} \\
                    &                                                     & Multi-property Optimization & 2,810 & MuMOInstruct \cite{mumoinstruct} \\ \hline
    \multicolumn{3}{c}{\textbf{Total}} & \multicolumn{2}{|c}{\textbf{$58,904$}} \\ \hline

    \end{tabular}
    \label{tab:bench}
\end{table}

For molecular understanding, we include all four name conversion tasks (I2S, I2F, S2I, and S2F), the molecule captioning task, and four property prediction tasks (BBBP-cls, ClinTox-cls, ESOL-reg, and LIPO-reg). For molecular generation, we cover two chemical reaction tasks (forward synthesis and retrosynthesis), four molecule editing tasks (add component, delete component, substitute component, and description-guided generation), and four property optimization tasks.

Specifically, we select five multi-property optimization tasks from MuMOInstruct \cite{mumoinstruct}, which are independent of a specific target:
\begin{itemize}
    \item BPQ (BBBP \& plogP \& QED) Joint Optimization: This task prioritizes brain permeability and appropriate lipophilicity while ensuring the optimized molecules retain favorable drug-like properties.
    \item MPQ (Mutagenicity \& plogP \& QED) Joint Optimization: This scenario represents early-stage lead optimization, aiming to reduce genotoxic risks while ensuring adequate lipophilicity and drug-like properties.
    \item BHMQ (BBBP \& HIA \& Mutagenicity \& QED) Joint Optimization: Particularly relevant for orally administered central nervous system (CNS) drugs, this task emphasizes both brain and intestinal absorption as critical factors.
    \item BMPQ (BBBP \& Mutagenicity \& plogP \& QED) Joint Optimization: This task reflects CNS drug design by balancing adequate lipophilicity, reduced toxicity, and favorable drug-like properties, simulating realistic requirements for CNS-active drugs.
    \item HMPQ (HIA \& Mutagenicity \& plogP \& QED) Joint Optimization: This task represents optimization for orally administered drugs, focusing on absorption, genotoxic risk reduction, and overall drug-like quality.
\end{itemize}

\subsection{Evaluation Metrics}
We adopt the original evaluation metrics for each task , following \cite{llasmol, tomg-bench, mumoinstruct}.

\textbf{Name Conversion Tasks.} 
The exact match (EM) metric is adopted to measure whether the predictions exactly match the gold standards. We apply the following criteria: (1) SMILES strings are parsed into molecules, and they are considered a match only if the two molecules are identical; (2) Molecular formulas are determined to be a match if the set of atoms is the same and the corresponding numbers of each atom are identical; (3) For IUPAC names, the set composed of multiple parts separated by semicolons is compared to determine a match.

\textbf{Molecule Captioning Task.} 
As a natural language generation task, we employ text-based metrics to measure similarity, including METEOR, BLEU-2, BLEU-4, ROUGE-1, ROUGE-2, and ROUGE-L.

\textbf{Property Prediction Tasks.} 
We calculate the accuracy (Acc.) for binary classification tasks and use the root mean square error (RMSE) to measure the differences between predicted and actual values for regression tasks.

\textbf{Chemical Reaction Tasks.} 
To evaluate the generated molecules, we use three metrics: (1) Exact match (EM) indicates the proportion of predicted molecules that exactly match the gold standards; (2) Fingerprint Tanimoto Similarity (FTS) quantifies the structural similarities using the Tanimoto similarity of their Morgan fingerprints; (3) Validity (Valid) measures the ratio of valid molecules generated, following SMILES grammar and chemical valence rules.

\textbf{Molecule Editing.} 
We employ the EM, FTS, and Valid metrics to measure the performance in the description-guided generation task. For component editing tasks, we adopt the success rate (SR), which is the percentage of molecules that successfully pass the rigorous testing processes \cite{tomg-bench}. The testing process, developed with RDKit, compares the generated molecule with the original one and verify whether the number of specified functional groups is correct after the required modification.

\textbf{Property Optimization.} 
We also adopt the success rate (SR) as a metric to indicate the proportion of generated molecules for which the specified property or all desired properties are successfully optimized as required. For each multi-property optimization instruction, we generate $20$ molecules via beam search decoding, following the evaluation setting in \cite{mumoinstruct}.

\section{Experiments}
We evaluate BioMedGPT-Mol on molecular understanding and generation tasks and compare its performance with other models. BioMedGPT-Mol shows remarkable performance in molecular discovery scenarios.

\subsection{Molecular Understanding}
Firstly, we evaluate the model's capabilities in component recognition, specific property prediction, and profile generation.

Across all LLMs, BioMedGPT-Mol achieves the best performance on name conversion (Table~\ref{tab:exp_name_conversion}), molecule captioning (Table~\ref{tab:exp_mol_cap}), and property prediction (Table~\ref{tab:exp_property_pred}). It not only surpasses general-purpose LLMs (GPT-4 \cite{gpt4}, Claude-3 Opus \cite{claude3}, and Galactica \cite{galactica}) by substantial margins, but also outperforms the three chemistry LLMs (MolInst \cite{mol-instructions}, ChemLLM \cite{chemllm}, and LlaSMol \cite{llasmol}), underscoring its mastery of molecular language.

On name conversion (Table~\ref{tab:exp_name_conversion}), BioMedGPT-Mol improves over LlaSMol by an absolute $7\%$ EM on average, evidencing its strong component-level understanding, which is the foundation of molecular understanding. For general LLMs, this task remains challenging; Claude-3 Opus attains only $15.4\%$ average EM. For reference, we follow \cite{llasmol} and include task-specific baselines: STOUT \cite{stout}, an encoder-decoder model specifically trained on SMILES–IUPAC pairs, handles I2S and S2I; a deterministic RDKit routine covers S2F; and I2F is decomposed into I2S+S2F. BioMedGPT-Mol has further narrowed the gap between LLMs and these specialized solutions.

\begin{table}[h]
    \centering
    \caption{\textbf{Performance on Name Conversion.} RDKit and STOUT \cite{stout} are used as SOTA task-specific models \cite{llasmol}. All the EM scores are in \%. The best performances of LLM-based models are highlighted in bold.}
    \begin{tabular}{c|c|cccc}
    \hline
    \bf Methods     & \bf Avg. & \bf I2F   &  \bf I2S   & \bf S2F   & \bf S2I \\ \hline
    \rowcolor{gray!20}
    \multicolumn{6}{c}{\textbf{Task Specific, Non-LLM based Models}} \\
    SOTA        & 82.0 & 97.9   & 73.5  & 100   & 56.5 \\ \hline
    \rowcolor{gray!20}
    \multicolumn{6}{c}{\textbf{General LLM}} \\
    GPT-4 \cite{gpt4} & 4.2 & 8.7 & 3.3 & 4.8 & 0.0 \\
    Galactica \cite{galactica} & 4.7 & 9.1 & 9.7 & 0.0 & 0.0 \\
    Claude3 Opus \cite{claude3} & 15.4 & 34.6 & 17.7 & 9.2 & 0.0 \\ \hline
    \rowcolor{gray!20}
    \multicolumn{6}{c}{\textbf{Chemistry LLM}} \\
    Molinst \cite{mol-instructions} & 0.0 & 0.0 & 0.0 & 0.0 & 0.0 \\
    ChemLLM \cite{chemllm} & 0.3 & 0.8 & 0.3 & 0.0 & 0.0 \\
    LlaSMol \cite{llasmol} & 70.1 & 87.9 & 70.1 & 93.2 & 29.0 \\ \hline
    BioMedGPT-Mol & \bf 77.1 & \bf 91.4 & \bf 76.5 & \bf 95.7 & \bf 44.9 \\ \hline                          
    \end{tabular}
    \label{tab:exp_name_conversion}
\end{table}

\begin{table}[ht]
    \centering
    \caption{\textbf{Performance on Molecule Captioning.} BL: BLEU. R: ROUGE. The best performances of LLM-based models are highlighted in bold.}
    \begin{tabular}{ccccccc}
    \hline
    \bf Methods & \bf METEOR & \bf BL-2 & \bf BL-4 & \bf R-1 & \bf R-2 & \bf R-L \\ \hline
    \rowcolor{gray!20}
    \multicolumn{7}{c}{\textbf{Task Specific, Non-LLM based Models}} \\
    Mol-T5 \cite{mol-t5}  & 0.515 & 0.462 & 0.366 & 0.563 & 0.398 & 0.501 \\ \hline
    \rowcolor{gray!20}
    \multicolumn{7}{c}{\textbf{General LLM}} \\
    Galactica \cite{galactica} & 0.050 & 0.018 & 0.002 & 0.061 & 0.012 & 0.052 \\
    GPT-4 \cite{gpt4} & 0.188 & 0.095 & 0.020 & 0.238 & 0.058 & 0.156 \\
    Claude3 Opus \cite{claude3} & 0.219 & 0.114 & 0.030 & 0.263 & 0.069 & 0.173 \\ \hline
    \rowcolor{gray!20}
    \multicolumn{7}{c}{\textbf{Chemistry LLM}} \\
    ChemLLM \cite{chemllm} & 0.050 & 0.012 & 0.004 & 0.057 & 0.005 & 0.048 \\
    Molinst \cite{mol-instructions} & 0.124 & 0.028 & 0.020 & 0.226 & 0.160 & 0.217 \\
    LlaSMol \cite{llasmol} & 0.452 & 0.414 & 0.319 & 0.521 & 0.357 & 0.463 \\ \hline
    BioMedGPT-Mol & \bf 0.515 & \bf 0.477 & \bf 0.381 & \bf 0.562 & \bf 0.403 & \bf 0.501 \\ \hline                          
    \end{tabular}
    \label{tab:exp_mol_cap}
\end{table}

\begin{table}[ht]
    \centering
    \caption{\textbf{Performance on Property Prediction.} The fine-tuned Uni-Mol \cite{unimol} is used as SOTA task-specific models \cite{llasmol}. All the accuracy scores are in \%. The best performances of LLM-based models are highlighted in bold.}
    \begin{tabular}{c|ccc|ccc}
    \hline
    \bf Methods    & \bf Avg.-reg & \bf ESOL   &  \bf LIPO & \bf Avg.-cls  & \bf BBBP   & \bf ClinTox \\ 
                   & RMSE $\downarrow$  & RMSE $\downarrow$ & RMSE $\downarrow$ & Acc. & Acc. & Acc. \\ \hline
    \rowcolor{gray!20}
    \multicolumn{7}{c}{\textbf{Task Specific, Non-LLM based Models}} \\
    Uni-Mol \cite{unimol} & 0.716 & 0.819  & 0.612 & 88.9  & 85.3   & 92.4 \\ \hline
    \rowcolor{gray!20}
    \multicolumn{7}{c}{\textbf{General LLM}} \\
    GPT-4 \cite{gpt4} & 2.058 & 2.570 & 1.545 & 56.5 & 62.9 & 50.0 \\
    Claude3 Opus \cite{claude3} & 1.115 & 1.036 & 1.194 & 58.4 & 75.1 & 41.7 \\
    Galactica \cite{galactica} & 3.582 & 4.184 & 2.979 & 80.7 & 69.0 & 92.4 \\ \hline
    \rowcolor{gray!20}
    \multicolumn{7}{c}{\textbf{Chemistry LLM}} \\
    Molinst \cite{mol-instructions} & 1.981 & 2.271 & 1.691 & 33.6 & 60.9 & 6.3 \\
    ChemLLM \cite{chemllm} & 1.872 & 1.946 & 1.797 & 49.0 & 22.3 & 75.7 \\
    LlaSMol \cite{llasmol} & 1.080 & 1.150 & 1.010 & 83.9 & 74.6 & \bf 93.1 \\ \hline
    BioMedGPT-Mol & \bf 0.945 & \bf 0.916 & \bf 0.973 & \bf 90.4 & \bf 87.6 & \bf 93.1 \\ \hline                          
    \end{tabular}
    \label{tab:exp_property_pred}
\end{table}

In molecule captioning (Table~\ref{tab:exp_mol_cap}), BioMedGPT-Mol records a METEOR score of $0.515$, outperforming LlaSMol \cite{llasmol} ($0.452$) and Claude-3 Opus \cite{claude3} ($0.219$). Compared with the task-specific Mol-T5 \cite{mol-t5} baseline, BioMedGPT-Mol delivers competitive performance.

Property prediction (Table~\ref{tab:exp_property_pred}) comprises both regression and classification tasks. For regression (ESOL and LIPO), BioMedGPT-Mol attains an average RMSE of $0.945$, demonstrating strong modeling of intrinsic properties. On classification, it reaches $90.4\%$ average accuracy, ranking first among all LLMs and even surpasses the task-specific baseline Uni-Mol \cite{unimol}. These experimental results underscore the model’s grasp of the intricate links among molecular components, structure, and properties.

\subsection{Molecular Generation}

Next, we assess the model's capabilities in three areas: intrinsic chemical law-based generation, detailed instruction-guided generation (description-guided generation and molecular component editing), and specific property objective-guided generation. Benefiting from its superior molecular understanding capabilities, BioMedGPT-Mol outperforms both general LLMs and chemistry-specific LLMs on these molecular generation tasks, as shown in Table~\ref{tab:exp_chem_reaction}, \ref{tab:exp_mol_gen}, \ref{tab:exp_mol_edit}, \ref{tab:exp_single_opt}, and \ref{tab:exp_multi_opt}.

\begin{table}[h]
    \centering
    \caption{\textbf{Performance on Chemical Reaction.} RSMILES \cite{RSMILES} and Molecular Transformer \cite{mol-trans} are re-trained as the SOTA task-specific models for forward synthesis and retrosynthesis respectively \cite{llasmol}. All the reported scores are in \%. The best performances of LLM-based models are highlighted in bold.}
    \begin{tabular}{c|c|cccccc}
    \hline
    \multirow{2}{*}{\bf Methods} & \bf Avg. & \multicolumn{3}{c}{\bf Forward Synthesis} & \multicolumn{3}{c}{\bf Retrosynthesis} \\
              & \bf EM & \bf EM & \bf FTS & \bf Valid & \bf EM & \bf FTS & \bf Valid \\ \hline
    \rowcolor{gray!20}
    \multicolumn{8}{c}{\textbf{Task Specific, Non-LLM based Models}} \\
    SOTA  & 62.9 & 78.7 & 92.2 & 100.0 & 47.0 & 77.5 & 99.7 \\ \hline
    \rowcolor{gray!20}
    \multicolumn{8}{c}{\textbf{General LLM}} \\
    Galactica \cite{galactica} & 0.0 & 0.0 & 25.9 & 83.7 & 0.0 & 34.6 & 93.0 \\
    GPT-4 \cite{gpt4} & 0.8 & 1.6 & 40.5 & 87.0 & 0.0 & 33.4 & 42.6 \\
    Claude3 Opus \cite{claude3} & 2.4 & 3.7 & 45.7 & 97.0 & 1.1 & 46.2 & 94.8 \\ \hline
    \rowcolor{gray!20}
    \multicolumn{8}{c}{\textbf{Chemistry LLM}} \\
    ChemLLM \cite{chemllm} & 0.0 & 0.0 & 1.6 & 38.5 & 0.0 & 2.9 & 10.9 \\
    Molinst \cite{mol-instructions} & 3.9 & 2.1 & 31.7 & \bf 99.8 & 5.7 & 48.0 & 97.8 \\
    LlaSMol \cite{llasmol} & 48.1 & 63.3 & 84.9 & \bf 99.8 & \bf 32.9 & 70.4 & \bf 100.0 \\ \hline
    BioMedGPT-Mol & \bf 49.8 & \bf 67.2 & \bf 88.5 & \bf 99.8 & 32.4 & \bf 75.2 & 99.9 \\ \hline                          
    \end{tabular}
    \label{tab:exp_chem_reaction}
\end{table}

We evaluate BioMedGPT-Mol on forward synthesis and retrosynthesis tasks (Table~\ref{tab:exp_chem_reaction}) to assess its understanding of intrinsic chemical laws, which underpins all molecular generation tasks. BioMedGPT-Mol achieves the best overall performance among all LLMs, with an average EM score of $49.8\%$. General LLMs struggle to capture in-domain rules, resulting in unsatisfactory performance. Even the two chemistry LLMs, Molinst \cite{mol-instructions} and ChemLLM \cite{chemllm}, face challenges with these tasks. LlaSMol \cite{llasmol} achieves an average EM score of $48.1\%$ and performs better on retrosynthesis. For reference, two transformer-based encoder-decoder models, RSMILES \cite{RSMILES} and Molecular Transformer \cite{mol-trans}, specifically adapted for these tasks, serve as task-specific baselines \cite{llasmol}. A performance gap persists between LLM-based approaches and specialized solutions. Further research is needed to enhance the capabilities of LLMs and narrow this gap.

\begin{table}[h]
    \centering
    \caption{\textbf{Performance on Description-guided Generation.} All the reported scores are in \%. The best performances of LLM-based models are highlighted in bold.}
    \begin{tabular}{cccc}
    \hline
    \bf Methods & \bf EM & \bf FTS & \bf Valid \\ \hline
    \rowcolor{gray!20}
    \multicolumn{4}{c}{\textbf{Task Specific, Non-LLM based Models}} \\
    Mol-T5 \cite{mol-t5}  & 31.7 & 73.2 & 95.3 \\ \hline
    \rowcolor{gray!20}
    \multicolumn{4}{c}{\textbf{General LLM}} \\
    Galactica \cite{galactica} & 0.0 & 11.6 & 94.7 \\
    GPT-4 \cite{gpt4} & 6.4 & 42.6 & 81.4 \\
    Claude3 Opus \cite{claude3} & 12.3 & 57.6 & 92.6 \\ \hline
    \rowcolor{gray!20}
    \multicolumn{4}{c}{\textbf{Chemistry LLM}} \\
    ChemLLM \cite{chemllm} & 0.9 & 14.3 & 4.3 \\
    Molinst \cite{mol-instructions} & 6.0 & 43.6 & 84.8 \\
    LlaSMol \cite{llasmol} & 19.2 & 61.7 & \bf 99.7 \\ \hline
    BioMedGPT-Mol & \bf 29.6 & \bf 77.5 & 98.9  \\ \hline                          
    \end{tabular}
    \label{tab:exp_mol_gen}
\end{table}

For description-guided molecular generation (Table~\ref{tab:exp_mol_gen}), models are tasked with generating a molecule that fits the given description. BioMedGPT-Mol achieves an EM score of $29.6\%$ and a FTS score of $77.5\%$, surpassing LlaSMol \cite{llasmol} by $10.4\%$ and $15.8\%$, respectively. Given the open-ended nature of this generation task, molecules that do not exactly match the gold standard may still be considered acceptable. BioMedGPT-Mol excels in FTS similarity score, even outperforming the task-specific baseline Mol-T5 \cite{mol-t5}. Experimental results highlight its strong capability in translating between natural language and molecular language. 

\begin{table}[h]
    \centering
    \caption{\textbf{Performance on Molecular Component Editing.} All the SR scores are in \%. The best performances of LLM-based models are highlighted in bold.}
    \begin{tabular}{c|c|ccc}
    \hline
    \bf Methods & \bf Avg. & \bf Add Comp. & \bf Delete Comp. & \bf Sub. Comp. \\ \hline
    \rowcolor{gray!20}
    \multicolumn{5}{c}{\textbf{Task Specific, Non-LLM based Models}} \\
    Bio-T5 \cite{biot5} & 19.4 & 34.6 & 16.7 & 6.8 \\
    Mol-T5 \cite{mol-t5}  & 22.5 & 28.3 & 22.3 & 16.9 \\ \hline
    \rowcolor{gray!20}
    \multicolumn{5}{c}{\textbf{General LLM}} \\
    Claude3.5 \cite{claude35} & 67.8 & 68.3 & 54.1 & \bf 81.0 \\
    GPT-4o \cite{gpt4} & 70.6 & 61.9 & 70.1 & 79.9 \\
    Gemini-1.5-pro \cite{gemini15pro} & 72.6 & 70.6 & 75.9 & 71.5 \\
    GPT-4-turbo \cite{gpt4} & 73.4 & 69.9 & 72.4 & 77.8 \\ \hline
    \rowcolor{gray!20}
    \multicolumn{5}{c}{\textbf{Chemistry LLM}} \\
    OpenMolIns-125M \cite{tomg-bench} & 47.5 & 58.4 & 65.3 & 18.7 \\
    OpenMolIns-8B \cite{tomg-bench} & 54.5 & 58.2 & 51.0 & 54.4 \\ \hline
    BioMedGPT-Mol & \bf 74.2 & \bf 72.9 & \bf 88.4 & 61.3 \\ \hline                          
    \end{tabular}
    \label{tab:exp_mol_edit}
\end{table}

Unlike generation from scratch, molecular component editing (Table~\ref{tab:exp_mol_edit}) demands a understanding of molecular structural components and editing instructions to execute modifications effectively. General LLMs show promising performance in this area. GPT-4 Turbo \cite{gpt4} stands out as the top performer, achieving an average success rate of $73.4\%$. However, the performance of the existing chemistry LLM, OpenMolIns \cite{tomg-bench}, highlights the considerable room for improvement. BioMedGPT-Mol achieves a remarkable success rate of $74.2\%$, surpassing GPT-4 Turbo. Despite these advancements, there is still room for improvement, especially in the substitution task, where Claude3.5 \cite{claude35} demonstrates a significant advantage.

\begin{table}[h]
    \centering
    \caption{\textbf{Performance on Single Property Optimization.} All the SR scores are in \%. The best performances of LLM-based models are highlighted in bold.}
    \begin{tabular}{c|c|ccc}
    \hline
    \bf Methods & \bf Avg. & \bf logP & \bf MR & \bf QED \\ \hline
    \rowcolor{gray!20}
    \multicolumn{5}{c}{\textbf{Task Specific, Non-LLM based Models}} \\
    Mol-T5 \cite{mol-t5}  & 44.6 & 42.4 & 45.0 & 46.5 \\
    Bio-T5 \cite{biot5} & 51.0 & 51.6 & 50.6 & 50.7 \\ \hline
    \rowcolor{gray!20}
    \multicolumn{5}{c}{\textbf{General LLM}} \\
    GPT-4o \cite{gpt4} & 60.0 & 71.9 & 68.6 & 39.5 \\
    GPT-4-turbo \cite{gpt4} & 63.3 & 76.6 & 73.9 & 39.5 \\
    Claude3.5 \cite{claude35} & 67.6 & 79.7 & 69.6 & 53.6 \\
    Gemini-1.5-pro \cite{gemini15pro} & 67.6 & 77.1 & 78.8 & 47.0 \\ \hline
    \rowcolor{gray!20}
    \multicolumn{5}{c}{\textbf{Chemistry LLM}} \\
    OpenMolIns-125M \cite{tomg-bench} & 67.6 & 73.6 & 71.2 & 57.9 \\
    OpenMolIns-8B \cite{tomg-bench} & 68.0 & 80.5 & 71.2 & 52.2 \\ \hline
    BioMedGPT-Mol & \bf 77.2 & \bf 87.9 & \bf 80.0 & \bf 63.8 \\ \hline                          
    \end{tabular}
    \label{tab:exp_single_opt}
\end{table}

Property optimization tasks (Table~\ref{tab:exp_single_opt}) require models to steer molecules toward desired optimization targets, necessitating a deeper understanding of the relationship between molecular structure and properties. BioMedGPT-Mol excels in all three subtasks, achieving an average success rate of $77.2\%$, which outperforms the second-best model, OpenMolIns-8B \cite{tomg-bench}, by $9.2\%$. The experimental results highlight BioMedGPT-Mol's ability to provide critical insights into the potential pharmacological attributes of molecules, thereby assisting chemists in identifying viable drug candidates.

\begin{table}[h]
    \centering
    \caption{\textbf{Performance on Multi-property Optimization.} All the SR scores are in \%. The best performances of LLM-based models are highlighted in bold.}
    \begin{tabular}{c|c|ccccc}
    \hline
    \bf Methods & \bf Avg. & \bf BPQ & \bf MPQ & \bf BHMQ & \bf BMPQ & \bf HMPQ \\ \hline
    \rowcolor{gray!20}
    \multicolumn{7}{c}{\textbf{General LLM (0-shot)}} \\
    Mistral \cite{mistral} & 14.8 & 15.8 & 11.2 & 12.7 & 12.6 & 21.9 \\
    GPT-4o \cite{gpt4} & 23.9 & 36.4 & 19.4 & 17.8 & 25.1 & 20.8 \\
    Claude3.5 \cite{claude35} & 49.8 & 56.0 & 17.4 & 39.0 & 44.5 & 38.5 \\ \hline
    \rowcolor{gray!20}
    \multicolumn{7}{c}{\textbf{General LLM (few-shot)}} \\
    GPT-4o \cite{gpt4} & 25.0 & 40.0 & 21.4 & 14.4 & 24.1 & 25.0 \\
    Mistral \cite{mistral} & 51.8 & 68.6 & 59.6 & 34.8 & 49.2 & 46.9 \\
    Claude3.5 \cite{claude35} & 59.6 & 76.8 & 50.6 & 39.0 & 44.5 & 38.5 \\ \hline
    \rowcolor{gray!20}
    \multicolumn{7}{c}{\textbf{Chemistry LLM}} \\
    ChemLLM \cite{chemllm} & 4.2 & 4.8 & 6.2 & 1.7 & 5.2 & 3.1 \\
    LlaSMol \cite{llasmol} & 66.8 & 86.0 & 76.4 & 53.4 & 64.9 & 53.1 \\
    GeLLM3O-mistral \cite{mumoinstruct} & 92.2 & 96.8 & \bf 95.2 & 86.4 & 91.1 & 91.7 \\
    GeLLM3O-llama \cite{mumoinstruct} & 95.0 & 95.0 & 93.6 & \bf 93.2 & \bf 95.3 & \bf 97.9 \\ \hline
    BioMedGPT-Mol & \bf 95.2 & \bf 99.2 & 94.6 & \bf 93.2 & 94.2 & 94.8 \\ \hline                          
    \end{tabular}
    \label{tab:exp_multi_opt}
\end{table}

Multi-property joint optimization further challenges models to improve all required properties simultaneously, necessitating the balancing of multiple trade-offs and conflicting objectives. BioMedGPT-Mol substantially outperforms general LLMs by a large margin, achieving a significant average improvement of $59.7\%$ in success rate over the best general LLM baseline, Claude 3.5 (5-shot) \cite{claude35}. Fine-tuned with specialized knowledge, the chemistry LLMs designed for property optimization, GeLLM3Os \cite{mumoinstruct}, exhibit remarkable performance with success rates of $92.2\%$ and $95.0\%$ for two different versions. BioMedGPT-Mol surpasses them in overall performance, achieving a success rate of $95.2\%$, which underscores its potential for applications in lead optimization.

\subsection{Performance Advantage of Explicit Reasoning on Molecular Component Editing}

\begin{figure}[ht]
\begin{center}
\includegraphics[width=0.95\linewidth]{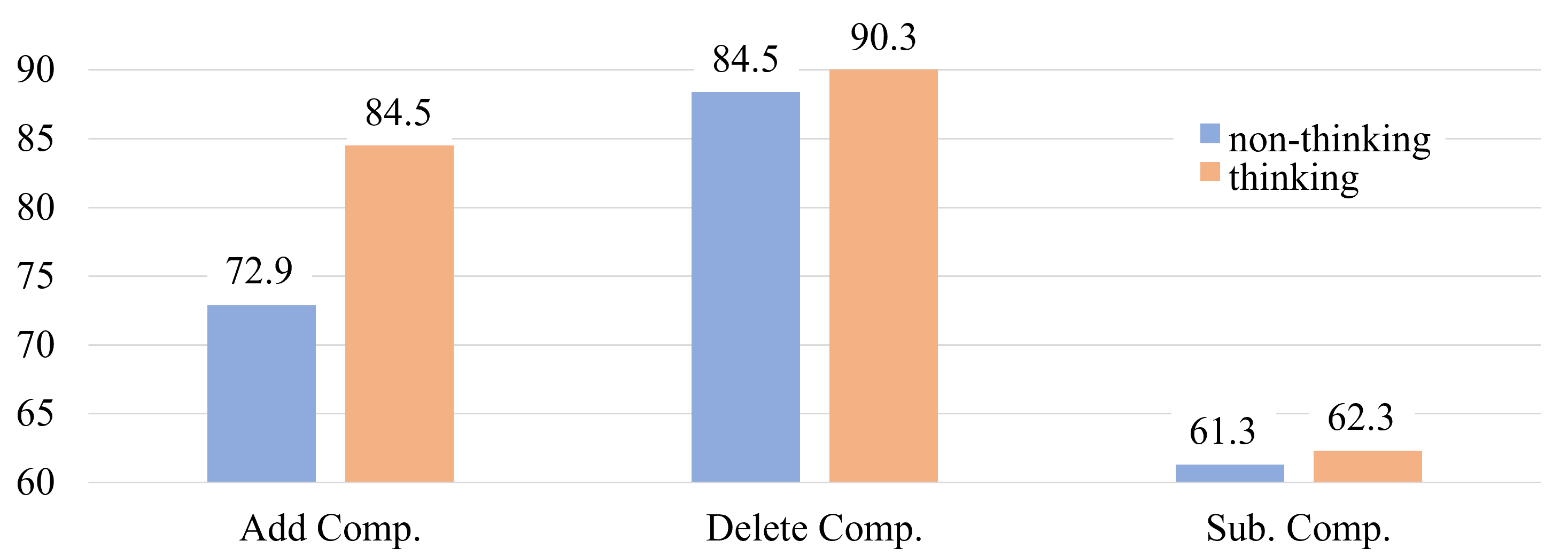}
\end{center}
   \caption{Performance comparison between non-thinking and thinking behavior on molecular component editing task.}
\label{fig:3}
\end{figure}

We further investigate a representative molecule-centric reasoning task, molecular component editing, to examine the performance differences between explicit and implicit reasoning behaviors, in other words, `thinking' and `non-thinking'. BioMedGPT-Mol allows users to manage and control reasoning behavior. We evaluate the model's performance on the editing task with explicit reasoning enabled. The performance comparison between the two reasoning behaviors is shown in Figure~\ref{fig:3}. Explicit reasoning behavior effectively improves performance across all three subtasks, resulting in an average success rate of $79.0\%$. The performance improvement on the `add component' task is particularly significant, with explicit reasoning outperforming implicit reasoning by $11.6\%$, setting a new state-of-the-art level. This highlights the potential power of molecular language models on reasoning tasks through increased inference computation budget.

\section{Further Exploring Retrosynthetic Planning with BioMedGPT-Mol}

Leveraging the power of BioMedGPT-Mol, we further explore retrosynthetic planning task.

Multi-step retrosynthetic planning is a fundamental problem in organic chemistry. It aims to identify a set of commercially available starting materials and a sequence of reactions to synthesize a given target molecule. Although recent works explore direct route generation to mitigate the exponential search space \cite{directmultistep}, existing deep learning-based methods predominantly employ iterative search algorithms \cite{segler2018planning, retro-star, dfpn-e, grasp, retro-graph, gnn-retro, hong2023retrosynthetic, yu2024doubleended}. Starting from the target, they iteratively select and deconstruct molecules until a viable set of starting materials is obtained. By integrating a policy network, Monte Carlo Tree Search (MCTS) effectively navigates the vast solution space, thereby significantly improving the efficiency of retrosynthetic planning \cite{segler2018planning, hong2023retrosynthetic}. Alternatively, Retro* \cite{retro-star} models the search problem as an AND-OR tree where AND nodes and OR nodes alternate, employing an A*-like heuristic to guide the search.

Among existing works, earlier studies adopted a context-free expansion policy (i.e., the single-step model) in which the context information of partially explored synthesis trees is ignored by the predictor \cite{segler2018planning, retro-star, dfpn-e, grasp, hong2023retrosynthetic}. More recently, studies argue that in-context reactions, namely the earlier predicted retrosynthetic steps, can be particularly beneficial to the expansion policy, regardless of the planner's framework. Algorithms like RetroGraph and GNN-Retro \cite{retro-graph, gnn-retro} utilize graph neural networks to aggregate information from the synthetic route, enabling context-aware cost estimation within the search heuristic. Building on this, FusionRetro \cite{liu2023fusionretro} explicitly models reactions along synthetic routes as reaction graphs, fusing context embeddings into the single-step predictor to improve planning performance. Furthermore, RetroInText \cite{kang2025retrointext} leverages multimodal LLMs for in-context representation learning, utilizing textual descriptions to guide intermediate identification. 

In contrast to these representation-focused approaches, our work investigates the potential of optimizing LLMs as standalone generative planners via explicit CoT reasoning. To the best of our knowledge, LLMs have not yet been successfully deployed as effective multi-step retrosynthetic planners in isolation. We demonstrate that BioMedGPT-Mol, empowered by its deep molecular expertise and optimized through SFT and RL, is capable of functioning as a standalone retrosynthetic planner, thereby bypassing the need for specialized external search algorithms.

In this section, we evaluate BioMedGPT-Mol as an end-to-end planner on the RetroBench dataset~\cite{liu2023fusionretro}. We adopt this benchmark for the rationality of its evaluation protocol, which employs a rigorous set-wise exact match metric. Unlike standard search success rates, this metric verifies whether the predicted starting materials can genuinely synthesize the target by checking for an exact match with the ground truth precursors. A prediction is deemed correct if it aligns with any feasible ground truth route. Furthermore, to ensure plausible route generation, the search is pruned if the predicted path length exceeds the depth of the ground truth pathway.

\subsection{A Three-stage SFT Strategy for Retrosynthetic Planning}

Building upon BioMedGPT-Mol, we employ a three-stage SFT strategy to adapt the model for the retrosynthetic planning task.

In the first stage, we fine-tune the model on the standard training set without CoT \cite{cot} to familiarize it with multi-step planning. For data augmentation, we extend the RSMILES technique \cite{RSMILES} to multi-step synthesis by aligning SMILES across all reactions and performing a 20-fold augmentation. Specifically, we randomly select a root atom in the target molecule and align the entire synthesis tree relative to that atom.

To enable explicit reasoning for multi-step planning, the second stage introduces a CoT training set. We distill DeepSeek-V3.1 \cite{ds-v3} using the complete retrosynthetic tree and all starting materials in SMILES format, prompting it to produce high-level retrosynthetic analyses. These analyses emphasize structural examination, identification of key reaction sites, and potential challenges across the synthesis plan, such as reaction ordering or the need for protecting groups.

Because natural-language reasoning is inherently more ambiguous than SMILES-based answer generation, it induces substantially higher uncertainty. Empirically, we observe that by the end of the second stage, the model’s per-token loss on the reasoning component is an order of magnitude higher than its loss on the answer component. This imbalance suggests that optimization is disproportionately influenced by the reasoning signal, potentially overshadowing supervision for the final prediction. Consequently, in the third stage, we rebalance the training objective to refocus the model on accurate reactant prediction. Specifically, within each training sample, we compute the average per-token cross-entropy loss for the reasoning and answer segments independently, denoted as $\mathcal{L}_{\text{thought}}$ and $\mathcal{L}_{\text{answer}}$, respectively. To ensure the model prioritizes the accuracy of the final retrosynthetic plan while maintaining logical coherence, we enforce a weighted objective function defined as:
\begin{equation}
    \mathcal{L} = \alpha \cdot \mathcal{L}_{\text{thought}} + (1-\alpha) \cdot \mathcal{L}_{\text{answer}}
\end{equation}
where $\alpha = 0.1$ in our implementation.
This weighting strategy effectively mitigates the impact of the higher entropy observed in the reasoning tokens, thereby stabilizing the optimization process.

\subsection{Reinforcement Learning with Verifiable Rewards}

Following the supervised fine-tuning stages, we further optimize BioMedGPT-Mol using RL via Group Relative Policy Optimization (GRPO)~\cite{deepseekmath}. The primary objective of this stage is to enable the model to generalize beyond the rigid trajectories found in the training data. Retrosynthetic planning is inherently one-to-many; while SFT teaches the model to mimic specific annotated routes, RL encourages the exploration of solution spaces. Our goal is for the model to identify the correct set of commercially available starting materials and generate chemically plausible intermediate steps, even if the planned route differs from the specific ground-truth path provided in RetroBench.

To guide this optimization, we design a hierarchical reward function comprising three progressive components. Crucially, a subsequent reward component is evaluated and awarded only if the preceding constraint is fully satisfied.

\begin{itemize}
    \item \textbf{Format Reward ($R_{\text{fmt}}$, weight=0.5):} This foundational reward ensures the model adheres to the required output structure, specifically verifying the correct usage of \texttt{<think>} tags for reasoning and the interpretable format of the answer. If the format is incorrect, the evaluation terminates, and subsequent rewards are set to zero.
    \item \textbf{Validity Reward ($R_{\text{valid}}$, weight=0.5):} Conditioned on a correct format, we verify the chemical validity of all generated molecular strings using RDKit. This reward penalizes the generation of invalid SMILES strings or chemically impossible structures.
    \item \textbf{Accuracy Reward ($R_{\text{acc}}$, weight=1.0):} Conditioned on both correct format and valid chemical syntax, this metric calculates the set-wise exact match between the predicted starting materials and the ground truth precursors. A positive reward is granted if the predicted set aligns with the feasible solution defined in RetroBench.
\end{itemize}

The total reward $R$ is the accumulated sum of these weighted components, effectively guiding the model from basic structural coherence to chemical validity, and finally to planning accuracy.

\subsection{Performance on RetroBench}
To investigate the performance of LLM in retrosynthetic planning task, we evaluate the fine-tuned model on RetroBench \cite{liu2023fusionretro} and compare it with other representative approaches, including template-based methods \cite{retrosym, neuralsym, gln}, semi-template-based methods \cite{g2gs, graphretro}, and template-based methods \cite{transformer-retro, megan, liu2023fusionretro, kang2025retrointext}.

\begin{figure}[ht]
\begin{center}
\includegraphics[width=0.8\linewidth]{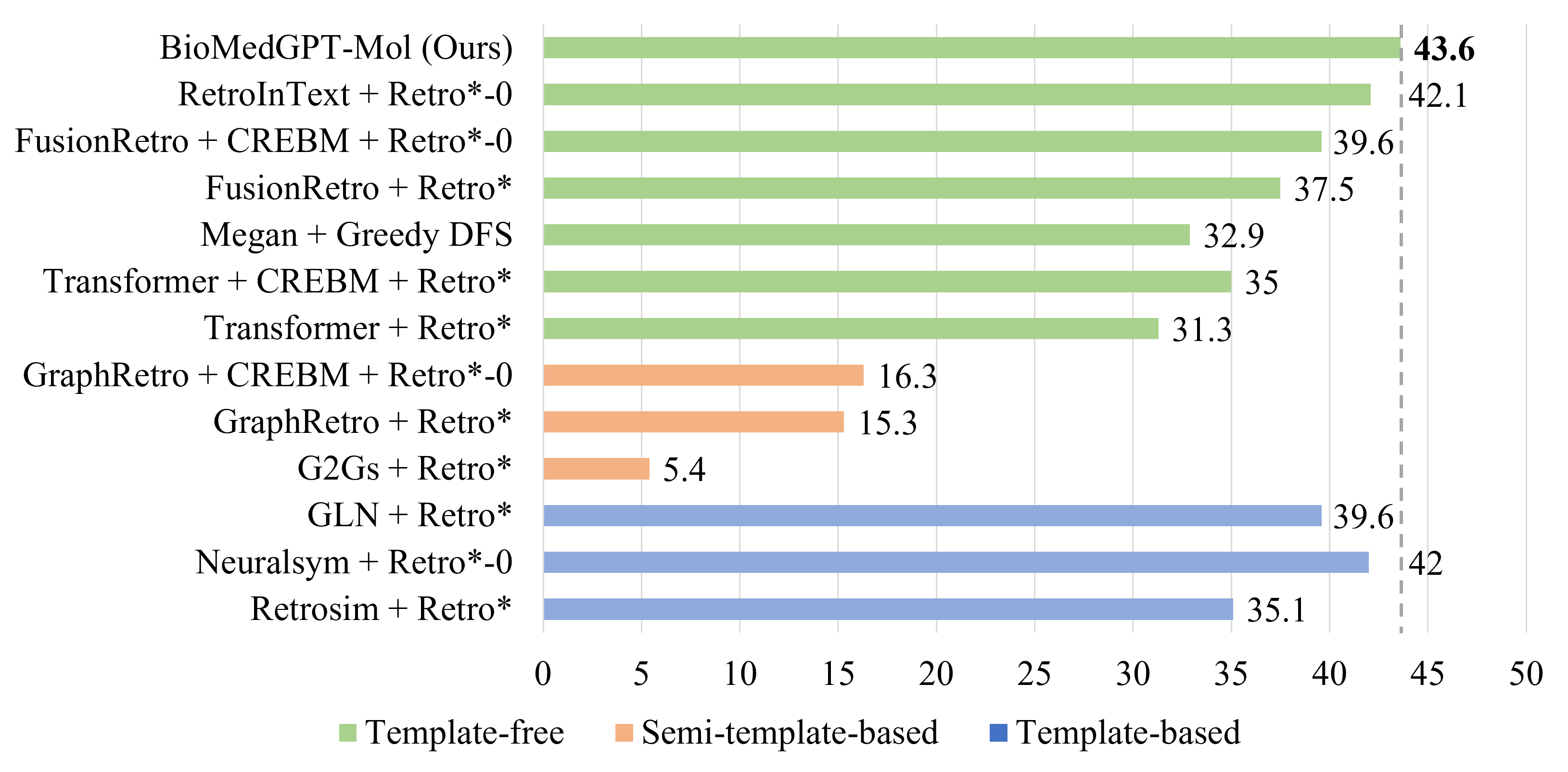}
\end{center}
   \caption{Summary of retrosynthetic planning results for Top-1 exact match accuracy (\%) on RetroBench.}
\label{fig:5}
\end{figure}

A two-dimensional test-time scaling strategy is adopted, incorporating SMILES augmentation and weighted beam search. We generate $k_a$ distinct SMILES representations for the target molecule by varying the root atoms and traversal orders. For each augmented input, we subsequently perform a partial beam search with a width of $k_b$ to decode the most probable reactant SMILES. The final predictions are ranked by aggregating the local beam scores across all augmented inputs using the reciprocal rank sum:
\begin{equation}
Score(y) = \sum_{j \in \mathcal{J}y} \frac{1}{r_{y,j}}
\end{equation}
where $y$ denotes a unique predicted reactant set, $\mathcal{J}_y$ comprises the indices of the augmented inputs where $y$ is generated, and $r_{y,j}$ represents the rank of candidate $y$ in the $j$-th partial beam search. In our experiments, we set $k_a = 6$ and $k_b = 2$.

Figure~\ref{fig:5} illustrates the exact-match accuracy ($\%$) across various retrosynthetic planning methods. Our model, BioMedGPT-Mol, achieves a state-of-the-art accuracy of $43.6\%$, outperforming all comparative baselines. Notably, our method surpasses both the strongest semi-template-based/template-free competitor, RetroInText ($42.1\%$), and the leading template-based model, NeuralSym ($42\%$). This superior performance underscores the efficacy of our fully end-to-end, template-free architecture, demonstrating its capability to accurately predict retrosynthetic routes without relying on rigid reaction templates or external search algorithms.

\section{Conclusion}

In this work, we introduce BioMedGPT-Mol, a molecular language model designed for molecular understanding and generation. To create a molecule-centric research assistant and harness the power of open-source reasoning models with rapid iterative upgrades, we design and implement a multi-task learning strategy for model fine-tuning. Our experimental results demonstrate its remarkable performance, showing that a general-purpose reasoning model can be effectively and efficiently post-trained into a specialized molecular language model through a well-structured multi-task curriculum. Building on this foundation, we further explored the challenging task of retrosynthetic planning. By implementing a comprehensive training strategy that integrates SFT and RL, we enabled BioMedGPT-Mol to achieve state-of-the-art performance as an end-to-end retrosynthetic planner.

\begin{figure}[ht]
\begin{center}
\includegraphics[width=0.8\linewidth]{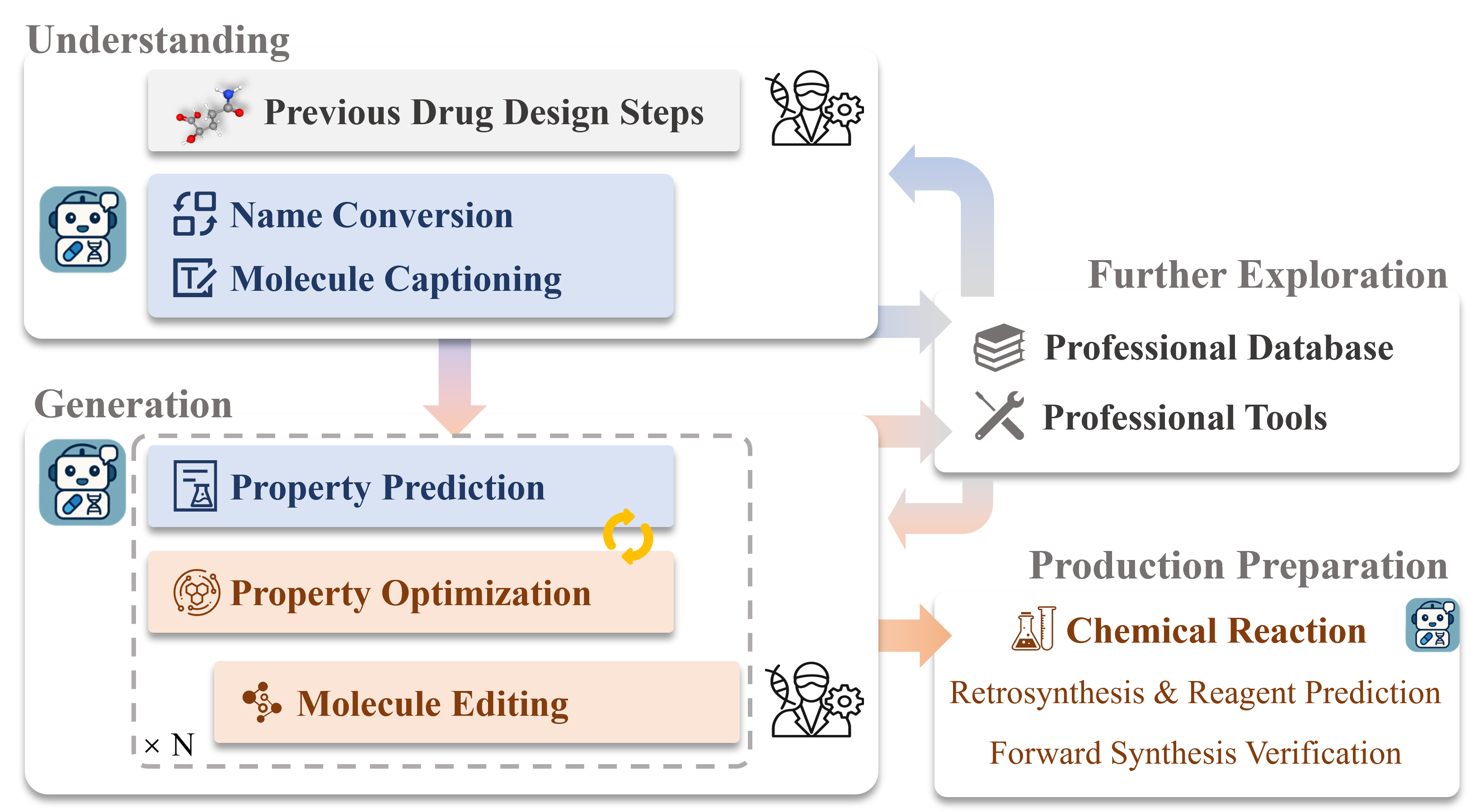}
\end{center}
   \caption{Potential workflow with BioMedGPT-Mol. BioMedGPT-Mol, with its capabilities in molecular understanding and generation, can support the entire molecular discovery journey, particularly in small molecule drug discovery. It can help experts more easily understand new molecules, optimize and generate satisfactory molecules, and propose methods for producing the generated molecules.}
\label{fig:4}
\end{figure}

BioMedGPT-Mol, with its capabilities in molecular understanding and generation, can assist the entire molecular discovery journey, especially for small molecule drug discovery (Figure~\ref{fig:4}). Experts can quickly get an overview of the input molecule through molecule captioning and explore further using external databases with the converted molecular names. When they focus on a specific molecule, the `prediction-optimization-editing' molecular generation cycle can help them more easily obtain a satisfactory molecule. Finally, BioMedGPT-Mol can propose a method for producing the output molecule. We anticipate that molecular language models, such as BioMedGPT-Mol, can significantly enhance expert-in-the-loop molecular research and development.

Furthermore, BioMedGPT-Mol can serve as a foundation for further molecule-centric model research, including optical chemical structure understanding, biomedical multimodal understanding and reasoning, and in-domain biomedical tool planning. Agent systems built with these models can facilitate molecule-centric scientific discovery in an autopilot manner.

{
    \small
    \bibliographystyle{ieeenat_fullname}
    \bibliography{main}

@String(ICLR = {Int. Conf. Learn. Represent.})

@String(AAAI = {AAAI})

@String(ICLR  = {ICLR})

@article{smiles,
  title={{SMILES}, a chemical language and information system. 1. Introduction to methodology and encoding rules},
  author={Weininger, David},
  journal={Journal of chemical information and computer sciences},
  volume={28},
  number={1},
  pages={31--36},
  year={1988}
}

@article{chemllm,
  title={{ChemLLM}: A chemical large language model},
  author={Zhang, Di and Liu, Wei and Tan, Qian and Chen, Jingdan and Yan, Hang and Yan, Yuliang and Li, Jiatong and Huang, Weiran and Yue, Xiangyu and Ouyang, Wanli and others},
  journal={arXiv preprint arXiv:2402.06852},
  year={2024}
}

@article{biomedgpt,
  title={{BioMedGPT}: Open multimodal generative pre-trained transformer for biomedicine},
  author={Luo, Yizhen and Zhang, Jiahuan and Fan, Siqi and Yang, Kai and Wu, Yushuai and Qiao, Mu and Nie, Zaiqing},
  journal={arXiv preprint arXiv:2308.09442},
  year={2023}
}

@article{ocsu,
  title={{OCSU}: Optical Chemical Structure Understanding for Molecule-centric Scientific Discovery},
  author={Fan, Siqi and Xie, Yuguang and Cai, Bowen and Xie, Ailin and Liu, Gaochao and Qiao, Mu and Xing, Jie and Nie, Zaiqing},
  journal={arXiv preprint arXiv:2501.15415},
  year={2025}
}

@article{tomg-bench,
  title={{TOMG-Bench}: Evaluating LLMs on text-based open molecule generation},
  author={Li, Jiatong and Li, Junxian and Liu, Yunqing and Zhou, Dongzhan and Li, Qing},
  journal={arXiv preprint arXiv:2412.14642},
  year={2024}
}

@article{qwen3,
  title={Qwen3 technical report},
  author={Yang, An and Li, Anfeng and Yang, Baosong and Zhang, Beichen and Hui, Binyuan and Zheng, Bo and Yu, Bowen and Gao, Chang and Huang, Chengen and Lv, Chenxu and others},
  journal={arXiv preprint arXiv:2505.09388},
  year={2025}
}

@article{ds-v3,
  title={Deepseek-v3 technical report},
  author={Liu, Aixin and Feng, Bei and Xue, Bing and Wang, Bingxuan and Wu, Bochao and Lu, Chengda and Zhao, Chenggang and Deng, Chengqi and Zhang, Chenyu and Ruan, Chong and others},
  journal={arXiv preprint arXiv:2412.19437},
  year={2024}
}

@misc{deepseekmath,
      title={DeepSeekMath: Pushing the Limits of Mathematical Reasoning in Open Language Models}, 
      author={Zhihong Shao and Peiyi Wang and Qihao Zhu and Runxin Xu and Junxiao Song and Xiao Bi and Haowei Zhang and Mingchuan Zhang and Y. K. Li and Y. Wu and Daya Guo},
      year={2024},
      eprint={2402.03300},
      archivePrefix={arXiv},
      primaryClass={cs.CL},
      url={https://arxiv.org/abs/2402.03300}, 
}

@article{ds-r1,
  title={Deepseek-r1: Incentivizing reasoning capability in llms via reinforcement learning},
  author={Guo, Daya and Yang, Dejian and Zhang, Haowei and Song, Junxiao and Zhang, Ruoyu and Xu, Runxin and Zhu, Qihao and Ma, Shirong and Wang, Peiyi and Bi, Xiao and others},
  journal={arXiv preprint arXiv:2501.12948},
  year={2025}
}

@article{o1,
  title={Openai o1 system card},
  author={Jaech, Aaron and Kalai, Adam and Lerer, Adam and Richardson, Adam and El-Kishky, Ahmed and Low, Aiden and Helyar, Alec and Madry, Aleksander and Beutel, Alex and Carney, Alex and others},
  journal={arXiv preprint arXiv:2412.16720},
  year={2024}
}

@article{kimi,
  title={Kimi k1. 5: Scaling reinforcement learning with llms},
  author={Team, Kimi and Du, Angang and Gao, Bofei and Xing, Bowei and Jiang, Changjiu and Chen, Cheng and Li, Cheng and Xiao, Chenjun and Du, Chenzhuang and Liao, Chonghua and others},
  journal={arXiv preprint arXiv:2501.12599},
  year={2025}
}

@article{deepscaler,
  title={Deepscaler: Surpassing o1-preview with a 1.5 b model by scaling rl},
  author={Luo, Michael and Tan, Sijun and Wong, Justin and Shi, Xiaoxiang and Tang, William Y and Roongta, Manan and Cai, Colin and Luo, Jeffrey and Zhang, Tianjun and Li, Li Erran and others},
  journal={Notion Blog},
  year={2025}
}

@article{llama-nemo,
  title={Llama-nemotron: Efficient reasoning models},
  author={Bercovich, Akhiad and Levy, Itay and Golan, Izik and Dabbah, Mohammad and El-Yaniv, Ran and Puny, Omri and Galil, Ido and Moshe, Zach and Ronen, Tomer and Nabwani, Najeeb and others},
  journal={arXiv preprint arXiv:2505.00949},
  year={2025}
}

@article{chempile,
  title={{ChemPile}: A 250GB Diverse and Curated Dataset for Chemical Foundation Models},
  author={Mirza, Adrian and Alampara, Nawaf and R{\'\i}os-Garc{\'\i}a, Marti{\~n}o and Abdelalim, Mohamed and Butler, Jack and Connolly, Bethany and Dogan, Tunca and Nezhurina, Marianna and {\c{S}}en, B{\"u}nyamin and Tirunagari, Santosh and others},
  journal={arXiv preprint arXiv:2505.12534},
  year={2025}
}

@article{ether0,
  title={Training a Scientific Reasoning Model for Chemistry},
  author={Narayanan, Siddharth M and Braza, James D and Griffiths, Ryan-Rhys and Bou, Albert and Wellawatte, Geemi and Ramos, Mayk Caldas and Mitchener, Ludovico and Rodriques, Samuel G and White, Andrew D},
  journal={arXiv preprint arXiv:2506.17238},
  year={2025}
}

@article{mol-instructions,
  title={Mol-instructions: A large-scale biomolecular instruction dataset for large language models},
  author={Fang, Yin and Liang, Xiaozhuan and Zhang, Ningyu and Liu, Kangwei and Huang, Rui and Chen, Zhuo and Fan, Xiaohui and Chen, Huajun},
  journal={arXiv preprint arXiv:2306.08018},
  year={2023}
}

@article{llasmol,
  title={{LlaSMol}: Advancing large language models for chemistry with a large-scale, comprehensive, high-quality instruction tuning dataset},
  author={Yu, Botao and Baker, Frazier N and Chen, Ziqi and Ning, Xia and Sun, Huan},
  journal={arXiv preprint arXiv:2402.09391},
  year={2024}
}

@article{mumoinstruct,
  title={$GeLLM^{3o}$: Generalizing Large Language Models for Multi-property Molecule Optimization},
  author={Dey, Vishal and Hu, Xiao and Ning, Xia},
  journal={arXiv preprint arXiv:2502.13398},
  year={2025}
}

@article{cotbench,
  title={Beyond Chemical QA: Evaluating LLM's Chemical Reasoning with Modular Chemical Operations},
  author={Li, Hao and Cao, He and Feng, Bin and Shao, Yanjun and Tang, Xiangru and Yan, Zhiyuan and Yuan, Li and Tian, Yonghong and Li, Yu},
  journal={arXiv preprint arXiv:2505.21318},
  year={2025}
}

@article{drugassist,
  title={{DrugAssist}: A large language model for molecule optimization},
  author={Ye, Geyan and Cai, Xibao and Lai, Houtim and Wang, Xing and Huang, Junhong and Wang, Longyue and Liu, Wei and Zeng, Xiangxiang},
  journal={Briefings in Bioinformatics},
  volume={26},
  number={1},
  pages={bbae693},
  year={2025},
  publisher={Oxford University Press}
}

@article{white2023assessment,
  title={Assessment of chemistry knowledge in large language models that generate code},
  author={White, Andrew D and Hocky, Glen M and Gandhi, Heta A and Ansari, Mehrad and Cox, Sam and Wellawatte, Geemi P and Sasmal, Subarna and Yang, Ziyue and Liu, Kangxin and Singh, Yuvraj and others},
  journal={Digital Discovery},
  volume={2},
  number={2},
  pages={368--376},
  year={2023},
  publisher={Royal Society of Chemistry}
}

@article{guo2023can,
  title={What can large language models do in chemistry? a comprehensive benchmark on eight tasks},
  author={Guo, Taicheng and Nan, Bozhao and Liang, Zhenwen and Guo, Zhichun and Chawla, Nitesh and Wiest, Olaf and Zhang, Xiangliang and others},
  journal={Advances in Neural Information Processing Systems},
  volume={36},
  pages={59662--59688},
  year={2023}
}

@article{jablonka2024leveraging,
  title={Leveraging large language models for predictive chemistry},
  author={Jablonka, Kevin Maik and Schwaller, Philippe and Ortega-Guerrero, Andres and Smit, Berend},
  journal={Nature Machine Intelligence},
  volume={6},
  number={2},
  pages={161--169},
  year={2024},
  publisher={Nature Publishing Group UK London}
}

@article{lm24,
  title={{L+M-24}: Building a dataset for language+ molecules@ acl 2024},
  author={Edwards, Carl and Wang, Qingyun and Zhao, Lawrence and Ji, Heng},
  journal={arXiv preprint arXiv:2403.00791},
  year={2024}
}

@article{moleculenet,
  title={{MoleculeNet}: a benchmark for molecular machine learning},
  author={Wu, Zhenqin and Ramsundar, Bharath and Feinberg, Evan N and Gomes, Joseph and Geniesse, Caleb and Pappu, Aneesh S and Leswing, Karl and Pande, Vijay},
  journal={Chemical science},
  volume={9},
  number={2},
  pages={513--530},
  year={2018}
}

@article{clintox,
  title={A data-driven approach to predicting successes and failures of clinical trials},
  author={Gayvert, Kaitlyn M and Madhukar, Neel S and Elemento, Olivier},
  journal={Cell chemical biology},
  volume={23},
  number={10},
  pages={1294--1301},
  year={2016}
}

@article{bbbp,
  title={A Bayesian approach to in silico blood-brain barrier penetration modeling},
  author={Martins, Ines Filipa and Teixeira, Ana L and Pinheiro, Luis and Falcao, Andre O},
  journal={Journal of chemical information and modeling},
  volume={52},
  number={6},
  pages={1686--1697},
  year={2012}
}

@article{esol,
  title={FreeSolv: a database of experimental and calculated hydration free energies, with input files},
  author={Mobley, David L and Guthrie, J Peter},
  journal={Journal of computer-aided molecular design},
  volume={28},
  number={7},
  pages={711--720},
  year={2014}
}

@article{lipo,
  title={Separation methods for estimating octanol--water partition coefficients},
  author={Poole, Salwa K and Poole, Colin F},
  journal={Journal of Chromatography B},
  volume={797},
  number={1-2},
  pages={3--19},
  year={2003},
  publisher={Elsevier}
}

@article{gpt4,
  title={{GPT-4} technical report},
  author={Achiam, Josh and Adler, Steven and Agarwal, Sandhini and Ahmad, Lama and Akkaya, Ilge and Aleman, Florencia Leoni and Almeida, Diogo and Altenschmidt, Janko and Altman, Sam and Anadkat, Shyamal and others},
  journal={arXiv preprint arXiv:2303.08774},
  year={2023}
}

@article{claude3,
  title={The claude 3 model family: Opus, sonnet, haiku},
  author={Anthropic},
  journal={Anthropic},
  year={2024}
}

@article{claude35,
  title={Claude-3.5},
  author={Anthropic},
  journal={Anthropic},
  year={2024}
}

@article{mistral,
  title={Mistral 7B},
  author={Mistral AI},
  journal={Mistral AI},
  year={2023}
}

@article{gemini15pro,
  title={Gemini-1.5-pro},
  author={Google Deepmind},
  journal={Google Deepmind},
  year={2024}
}

@article{galactica,
  title={Galactica: A large language model for science},
  author={Taylor, Ross and Kardas, Marcin and Cucurull, Guillem and Scialom, Thomas and Hartshorn, Anthony and Saravia, Elvis and Poulton, Andrew and Kerkez, Viktor and Stojnic, Robert},
  journal={arXiv preprint arXiv:2211.09085},
  year={2022}
}

@article{stout,
  title={{STOUT}: SMILES to IUPAC names using neural machine translation},
  author={Rajan, Kohulan and Zielesny, Achim and Steinbeck, Christoph},
  journal={Journal of Cheminformatics},
  volume={13},
  number={1},
  pages={34},
  year={2021},
  publisher={Springer}
}

@article{unimol,
  title={{Uni-Mol}: A universal 3d molecular representation learning framework},
  author={Zhou, Gengmo and Gao, Zhifeng and Ding, Qiankun and Zheng, Hang and Xu, Hongteng and Wei, Zhewei and Zhang, Linfeng and Ke, Guolin},
  journal={ICLR},
  year={2023}
}

@article{mol-t5,
  title={Translation between molecules and natural language},
  author={Edwards, Carl and Lai, Tuan and Ros, Kevin and Honke, Garrett and Cho, Kyunghyun and Ji, Heng},
  journal={arXiv preprint arXiv:2204.11817},
  year={2022}
}

@article{RSMILES,
  title={Root-aligned SMILES: a tight representation for chemical reaction prediction},
  author={Zhong, Zipeng and Song, Jie and Feng, Zunlei and Liu, Tiantao and Jia, Lingxiang and Yao, Shaolun and Wu, Min and Hou, Tingjun and Song, Mingli},
  journal={Chemical Science},
  volume={13},
  number={31},
  pages={9023--9034},
  year={2022},
  publisher={Royal Society of Chemistry}
}

@article{mol-trans,
  title={Molecular transformer: a model for uncertainty-calibrated chemical reaction prediction},
  author={Schwaller, Philippe and Laino, Teodoro and Gaudin, Th{\'e}ophile and Bolgar, Peter and Hunter, Christopher A and Bekas, Costas and Lee, Alpha A},
  journal={ACS central science},
  volume={5},
  number={9},
  pages={1572--1583},
  year={2019},
  publisher={ACS Publications}
}

@article{biot5,
  title={{BioT5}: Enriching cross-modal integration in biology with chemical knowledge and natural language associations},
  author={Pei, Qizhi and Zhang, Wei and Zhu, Jinhua and Wu, Kehan and Gao, Kaiyuan and Wu, Lijun and Xia, Yingce and Yan, Rui},
  journal={arXiv preprint arXiv:2310.07276},
  year={2023}
}

@inproceedings{liu2023fusionretro,
  title={FusionRetro: Molecule Representation Fusion via In-Context Learning for Retrosynthetic Planning},
  author={Liu, Songtao and Tu, Zhengkai and Xu, Minkai and Zhang, Zuobai and Lin, Lu and Ying, Rex and Tang, Jian and Zhao, Peilin and Wu, Dinghao},
  booktitle={International Conference on Machine Learning},
  year={2023}
}

@inproceedings{cot,
 author = {Wei, Jason and Wang, Xuezhi and Schuurmans, Dale and Bosma, Maarten and ichter, brian and Xia, Fei and Chi, Ed and Le, Quoc V and Zhou, Denny},
 booktitle = {Advances in Neural Information Processing Systems},
 editor = {S. Koyejo and S. Mohamed and A. Agarwal and D. Belgrave and K. Cho and A. Oh},
 pages = {24824--24837},
 publisher = {Curran Associates, Inc.},
 title = {Chain-of-Thought Prompting Elicits Reasoning in Large Language Models},
 url = {https://proceedings.neurips.cc/paper_files/paper/2022/file/9d5609613524ecf4f15af0f7b31abca4-Paper-Conference.pdf},
 volume = {35},
 year = {2022}
}

@inproceedings{
kang2025retrointext,
title={RetroInText: A Multimodal Large Language Model Enhanced Framework for Retrosynthetic Planning via In-Context Representation Learning},
author={Chenglong Kang and Xiaoyi Liu and Fei Guo},
booktitle={The Thirteenth International Conference on Learning Representations},
year={2025},
url={https://openreview.net/forum?id=J6e4hurEKd}
}

@inproceedings{
graphretro,
title={Learning Graph Models for Retrosynthesis Prediction},
author={Vignesh Ram Somnath and Charlotte Bunne and Connor W. Coley and Andreas Krause and Regina Barzilay},
booktitle={Thirty-Fifth Conference on Neural Information Processing Systems},
year={2021},
url={https://openreview.net/forum?id=SnONpXZ_uQ_}
}

@inproceedings{gln,
  title={Retrosynthesis Prediction with Conditional Graph Logic Network},
  author={Dai, Hanjun and Li, Chengtao and Coley, Connor and Dai, Bo and Song, Le},
  booktitle={Advances in Neural Information Processing Systems},
  pages={8870--8880},
  year={2019}
}

@article{neuralsym,
  title={Neural-symbolic machine learning for retrosynthesis and reaction prediction},
  author={Segler, Marwin HS and Waller, Mark P},
  journal={Chemistry--A European Journal},
  volume={23},
  number={25},
  pages={5966--5971},
  year={2017},
  publisher={Wiley Online Library}
}

@article{retrosym,
title = {Computer-aided retrosynthetic design: fundamentals, tools, and outlook},
journal = {Current Opinion in Chemical Engineering},
volume = {35},
pages = {100721},
year = {2022},
issn = {2211-3398},
doi = {https://doi.org/10.1016/j.coche.2021.100721},
url = {https://www.sciencedirect.com/science/article/pii/S2211339821000538},
author = {Yijia Sun and Nikolaos V Sahinidis}
}

@article{megan,
author = {Sacha, Mikołaj and Błaż, Mikołaj and Byrski, Piotr and Dąbrowski-Tumański, Paweł and Chromiński, Mikołaj and Loska, Rafał and Włodarczyk-Pruszyński, Paweł and Jastrzębski, Stanisław},
title = {Molecule Edit Graph Attention Network: Modeling Chemical Reactions as Sequences of Graph Edits},
journal = {Journal of Chemical Information and Modeling},
volume = {61},
number = {7},
pages = {3273-3284},
year = {2021},
doi = {10.1021/acs.jcim.1c00537},
    note ={PMID: 34251814},
URL = { 
        https://doi.org/10.1021/acs.jcim.1c00537
},
eprint = { 
        https://doi.org/10.1021/acs.jcim.1c00537
}
}

@inproceedings{g2gs,
  title={A Graph to Graphs Framework for Retrosynthesis Prediction},
  author={Chence Shi and Minkai Xu and Hongyu Guo and Ming Zhang and Jian Tang},
  booktitle={International Conference on Machine Learning},
  year={2020},
  url={https://api.semanticscholar.org/CorpusID:214713971}
}

@inproceedings{transformer-retro,
author = {Karpov, Pavel and Godin, Guillaume and Tetko, Igor V.},
title = {A Transformer Model for Retrosynthesis},
year = {2019},
isbn = {978-3-030-30492-8},
publisher = {Springer-Verlag},
address = {Berlin, Heidelberg},
url = {https://doi.org/10.1007/978-3-030-30493-5_78},
doi = {10.1007/978-3-030-30493-5_78},
booktitle = {Artificial Neural Networks and Machine Learning – ICANN 2019: Workshop and Special Sessions: 28th International Conference on Artificial Neural Networks, Munich, Germany, September 17–19, 2019, Proceedings},
pages = {817–830},
numpages = {14},
location = {Munich, Germany}
}

@inproceedings{retro-star,
author = {Chen, Binghong and Li, Chengtao and Dai, Hanjun and Song, Le},
title = {Retro*: learning retrosynthetic planning with neural guided a* search},
year = {2020},
publisher = {JMLR.org},
booktitle = {Proceedings of the 37th International Conference on Machine Learning},
articleno = {150},
numpages = {9},
series = {ICML'20}
}

@article{gnn-retro, title={GNN-Retro: Retrosynthetic Planning with Graph Neural Networks}, volume={36}, url={https://ojs.aaai.org/index.php/AAAI/article/view/20318}, DOI={10.1609/aaai.v36i4.20318}, abstractNote={Retrosynthetic planning plays an important role in the field of organic chemistry, which could generate a synthetic route for the target product. The synthetic route is a series of reactions which are started from the available molecules. The most challenging problem in the generation of the synthetic route is the large search space of the candidate reactions. Estimating the cost of candidate reactions has been proved effectively to prune the search space, which could achieve a higher accuracy with the same search iteration. And the estimation of one reaction is comprised of the estimations of all its reactants. So, how to estimate the cost of these reactants will directly influence the quality of results. To get a better performance, we propose a new framework, named GNN-Retro, for retrosynthetic planning problem by combining graph neural networks(GNN) and the latest search algorithm. The structure of GNN in our framework could incorporate the information of neighboring molecules, which will improve the estimation accuracy of our framework. The experiments on the USPTO dataset show that our framework could outperform the state-of-the-art methods with a large margin under the same settings.}, number={4}, journal={Proceedings of the AAAI Conference on Artificial Intelligence}, author={Han, Peng and Zhao, Peilin and Lu, Chan and Huang, Junzhou and Wu, Jiaxiang and Shang, Shuo and Yao, Bin and Zhang, Xiangliang}, year={2022}, month={Jun.}, pages={4014-4021} }

@article{segler2018planning,
  title={Planning chemical syntheses with deep neural networks and symbolic {AI}},
  author={Segler, Marwin H. S. and Preuss, Mike and Waller, Mark P.},
  journal={Nature},
  volume={555},
  number={7698},
  pages={604--610},
  year={2018},
  publisher={Nature Publishing Group},
  doi={10.1038/nature25978},
  url={https://doi.org/10.1038/nature25978}
}

@article{hong2023retrosynthetic,
  title={Retrosynthetic planning with experience-guided {Monte} {Carlo} tree search},
  author={Hong, Siqi and Zhuo, Hankz Hankui and Jin, Kebing and Shao, Guang and Zhou, Zhanwen},
  journal={Communications Chemistry},
  volume={6},
  number={1},
  pages={120},
  year={2023},
  publisher={Nature Portfolio},
  doi={10.1038/s42004-023-00911-8},
  url={https://doi.org/10.1038/s42004-023-00911-8}
}

@inproceedings{grasp,
 author = {Yu, Yemin and Wei, Ying and Kuang, Kun and Huang, Zhengxing and Yao, Huaxiu and Wu, Fei},
 booktitle = {Advances in Neural Information Processing Systems},
 editor = {S. Koyejo and S. Mohamed and A. Agarwal and D. Belgrave and K. Cho and A. Oh},
 pages = {10257--10268},
 publisher = {Curran Associates, Inc.},
 title = {GRASP: Navigating Retrosynthetic Planning with Goal-driven Policy},
 url = {https://proceedings.neurips.cc/paper_files/paper/2022/file/42beaab8aa8da1c77581609a61eced93-Paper-Conference.pdf},
 volume = {35},
 year = {2022}
}

@inbook{dfpn-e,
author = {Kishimoto, Akihiro and Buesser, Beat and Chen, Bei and Botea, Adi},
title = {Depth-first proof-number search with heuristic edge cost and application to chemical synthesis planning},
year = {2019},
publisher = {Curran Associates Inc.},
address = {Red Hook, NY, USA},
booktitle = {Proceedings of the 33rd International Conference on Neural Information Processing Systems},
articleno = {649},
numpages = {11}
}

@inproceedings{retro-graph, series={KDD ’22},
   title={RetroGraph: Retrosynthetic Planning with Graph Search},
   url={http://dx.doi.org/10.1145/3534678.3539446},
   DOI={10.1145/3534678.3539446},
   booktitle={Proceedings of the 28th ACM SIGKDD Conference on Knowledge Discovery and Data Mining},
   publisher={ACM},
   author={Xie, Shufang and Yan, Rui and Han, Peng and Xia, Yingce and Wu, Lijun and Guo, Chenjuan and Yang, Bin and Qin, Tao},
   year={2022},
   month=aug, pages={2120–2129},
   collection={KDD ’22}
}

@inproceedings{yu2024doubleended,
 author = {Yu, Kevin and Roh, Jihye and Li, Ziang and Gao, Wenhao and Wang, Runzhong and Coley, Connor W.},
 booktitle = {Advances in Neural Information Processing Systems},
 doi = {10.52202/079017-3588},
 editor = {A. Globerson and L. Mackey and D. Belgrave and A. Fan and U. Paquet and J. Tomczak and C. Zhang},
 pages = {112919--112949},
 publisher = {Curran Associates, Inc.},
 title = {Double-Ended Synthesis Planning with Goal-Constrained Bidirectional Search},
 url = {https://proceedings.neurips.cc/paper_files/paper/2024/file/cd091a4d8e97157d32940428f902c7b0-Paper-Conference.pdf},
 volume = {37},
 year = {2024}
}

@article{directmultistep,
  author    = {Shee, Yu and Morgunov, Anton and Li, Haote and Batista, Victor S.},
  title     = {DirectMultiStep: Direct Route Generation for Multistep Retrosynthesis},
  journal   = {Journal of Chemical Information and Modeling},
  year      = {2025},
  volume    = {65},
  number    = {8},
  pages     = {3903--3914},
  doi       = {10.1021/acs.jcim.4c01982},
  url       = {https://doi.org/10.1021/acs.jcim.4c01982},
  publisher = {American Chemical Society},
  issn      = {1549-9596}
}
}




\end{document}